\documentclass{article} % For LaTeX2e

\makeatletter
\let\CiteVQA@addcontentsline\addcontentsline
\makeatother

\usepackage{iclr2027_conference,times}
\iclrfinalcopy % Uncomment for camera-ready version, but NOT for submission.

\makeatletter
\let\addcontentsline\CiteVQA@addcontentsline
\makeatother

\usepackage[utf8]{inputenc}
\usepackage[T1]{fontenc}
\usepackage{microtype}
\usepackage{xspace}

\usepackage[table]{xcolor}

\usepackage{tabulary}
\usepackage{etoolbox}

\newcommand{\highlightgreen}[1]{\colorbox[HTML]{d3ff9f}{\textbf{#1}}}

\newcommand{\highlightblue}[1]{\colorbox[HTML]{bae6fb}{\textbf{#1}}}

\usepackage{tcolorbox}
\tcbuselibrary{listings, breakable, skins}
\usepackage{listings}
\usepackage{tabularx}  
\usepackage{colortbl}
\usepackage{longtable}
\usepackage{booktabs}
\usepackage{multirow}
\usepackage{graphicx}
\usepackage{caption}
\usepackage{wrapfig}
\usepackage{setspace}
\usepackage{titletoc}
\usepackage{array}
\usepackage{makecell}
\usepackage{amsmath, amssymb, amsfonts}
\usepackage{algorithm}
\usepackage{algpseudocode}
\usepackage[misc]{ifsym}
\newcommand{\Jrel}{\mathcal{J}_{\text{rel}}}
\newcommand{\Jans}{\mathcal{J}_{\text{ans}}}
\usepackage{url}
\usepackage{hyperref}
\usepackage[nameinlink]{cleveref}

\definecolor{promptbackground}{RGB}{235, 245, 255}
\definecolor{promptframe}{RGB}{60, 120, 180}
\definecolor{outputbackground}{gray}{0.95}
\definecolor{outputframe}{gray}{0.65}

\newtcblisting{promptbox}[2][]{
    listing only, breakable, title=#2,
    colback=promptbackground, colframe=promptframe, coltitle=white,
    fonttitle=\bfseries, boxrule=0.5pt, arc=3mm, colbacktitle=promptframe,
    fontupper=\small\ttfamily,
    #1
}

\newcommand{\wzr}[1]{}
\newcommand{\yqh}[1]{}
\newcommand{\tip}[1]{}

\newcommand{\dataset}{CiteVQA} 

\makeatletter
\def\@maketitle{\vbox{\hsize\textwidth
{\LARGE\sc \@title\par}
\vskip 0.15in
{\centering \@author \par}
\vskip 0.3in minus 0.1in}}
\makeatother

\title{\dataset{}: Benchmarking Evidence Attribution \\for Trustworthy Document Intelligence}

\author{%
  {\bfseries
  Dongsheng Ma$^{1,2}$\thanks{Equal contribution.}\ , Jiayu Li$^{2}$\footnotemark[1]\ , Zhengren Wang$^{1,2}$\footnotemark[1]\ \\ Yijie Wang$^{2}$, Jiahao Kong$^{2}$, 
  Weijun Zeng$^{1,2}$, Jutao Xiao$^{2}$ \\Jie Yang$^{2}$, Bangrui Xu$^{2}$, Yuhan Wang$^{1}$, Bin Wang$^{2}$\thanks{Corresponding authors.}\ , Conghui He$^{2}$\footnotemark[2]} \\[6pt]
  $^{1}$Peking University \qquad $^{2}$Shanghai Artificial Intelligence Laboratory 
  \texttt{madongsheng26@stu.pku.edu.cn} \qquad \texttt{\{wangbin, heconghui\}@pjlab.org.cn}
}
\begin{document}

\maketitle
\lhead{Preprint. Under Review.}
\begin{abstract}
 
Multimodal Large Language Models (MLLMs) have significantly advanced document understanding, yet current Doc-VQA evaluations score only the final answer and leave the supporting evidence unchecked. This answer-only approach masks a critical failure mode: a model can land on the correct answer while grounding it in the wrong passage---a critical risk in high-stakes domains like law, finance, and medicine, where every conclusion must be traceable to a specific source region. To address this, we introduce \textbf{CiteVQA}, a benchmark that requires models to return \textit{element-level} bounding-box citations alongside each answer, evaluating both jointly. CiteVQA comprises 1,897 questions across 711 PDFs spanning seven domains and two languages, averaging 40.6 pages per document. To ensure fidelity and scalability, the ground-truth citations are generated by an automated pipeline---which identifies crucial evidence via masking ablation and enforces multi-stage quality control. At the core of our evaluation is Strict Attributed Accuracy (SAA), which credits a prediction only when the answer and the cited region are both correct. Auditing 20 MLLMs reveals a pervasive \textbf{Attribution Hallucination}: models frequently produce the right answer while citing the wrong region. The strongest system (Gemini-3.1-Pro-Preview) achieves an SAA of only 76.0, and the strongest open-source MLLM reaches just 22.5. Ultimately, towards trustworthy document intelligence, CiteVQA exposes a reliability gap that answer-only evaluations overlook, providing the instrumentation needed to close it.  Our repository is available at \url{https://github.com/opendatalab/CiteVQA}.

\end{abstract}

\section{Introduction}

In recent years, Multimodal Large Language Models (MLLMs) have achieved breakthrough progress in Document Understanding~\citep{ouyang2025omnidocbench}, demonstrating unprecedented capabilities in complex visual layout analysis and cross-modal reasoning. However, as model scale and performance escalate, a critical challenge has emerged: existing Document Visual Question Answering (Doc-VQA) evaluation frameworks focus almost exclusively on final answer accuracy~\citep{mathew2021docvqa,ma2024mmlongbench,tanaka2023slidevqa,mathew2022infographicvqa,wang2024charxiv,masry2022chartqa}, neglecting the logical path through which the model derives that answer---namely, the precise extraction of evidence. Consequently, the true depth and reliability of a model's comprehension remain largely unverified.

In high-stakes domains such as legal consultation, financial auditing, and evidence-based medicine, "evidence" is the cornerstone of decision-making~\citep{keer2026med,yu2025mramg}. An answer-only evaluation masks a critical failure mode: models might rely on pre-trained background knowledge to "make a guess," or land on the correct answer despite grounding it in the wrong passage. Such black-box reasoning poses uncontrollable risks of hallucination~\citep{wang2025rare,zhao2026retrieval}. Therefore, an urgent need exists for a benchmark that simultaneously evaluates answer accuracy and evidence faithfulness towards Trustworthy Document Intelligence, bridging the critical gap between text generation and source verification.

\begin{figure}[tb]
\centering
\includegraphics[width=\textwidth]{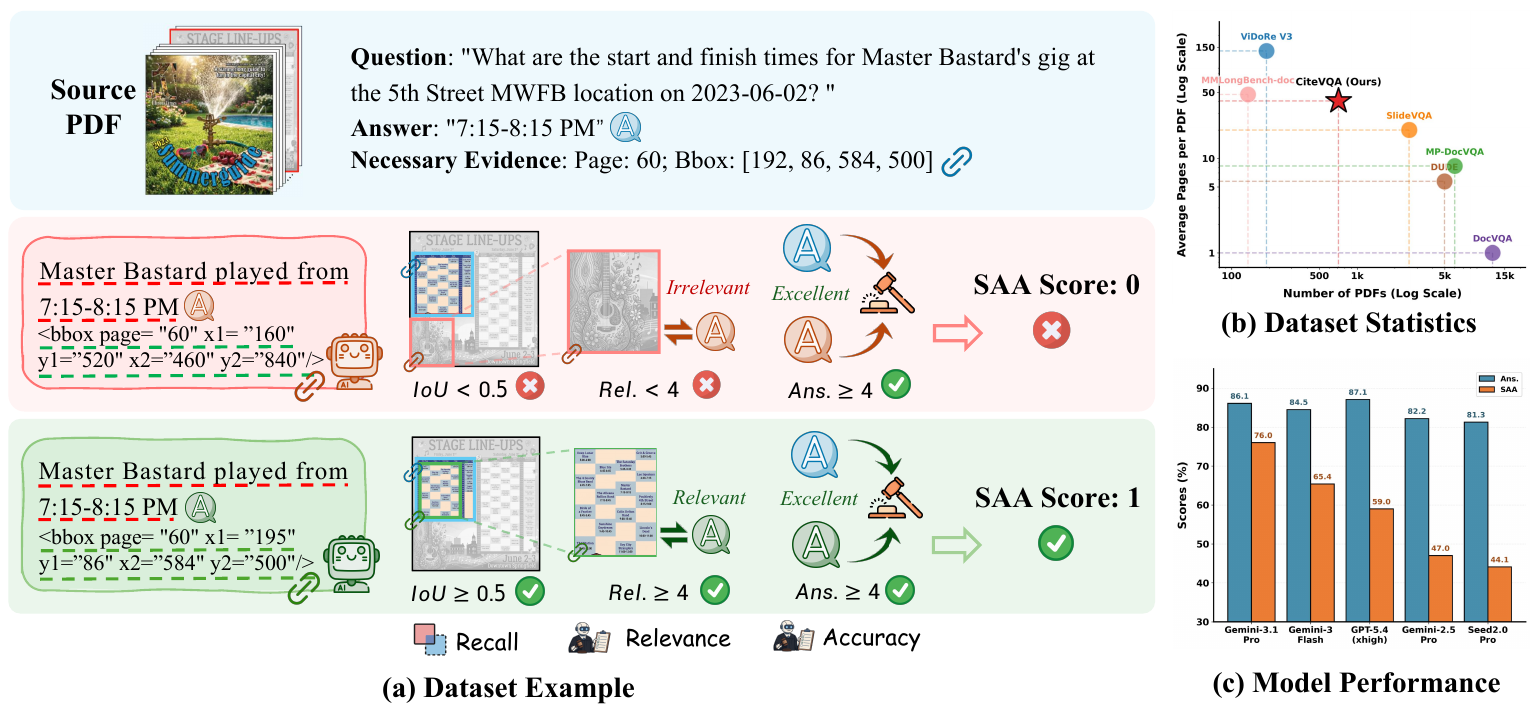}
\caption{Overview of the CiteVQA benchmark. (a) An example task requiring both correct answers and precise evidence citations to satisfy the Strict Attributed Accuracy (SAA) metric. (b) Dataset statistics: CiteVQA achieves a balance between document scale and page counts, better reflecting real-world complexity. (c) Performance of MLLMs: Despite high question accuracy, a significant gap exists in SAA due to "Attribution Hallucination".}
\label{fig:citevqa_example}
\end{figure}

To address these limitations, we introduce \textbf{CiteVQA}: A Benchmark for Faithful Evidence Attribution. Designed for long-form, multi-domain, and cross-lingual scenarios, CiteVQA comprises 1,897 high-quality questions derived from 711 PDFs across seven major domains. As illustrated in Figure~\ref{fig:citevqa_example}b, CiteVQA strikes a delicate balance between document quantity and length to better simulate real-world complexity. Unlike traditional tasks, CiteVQA mandates that models provide the precise PDF source supporting their answer at the granularity of \textit{element-level} bounding-box citations, thereby ensuring that every generated claim is visually verifiable by human users.

Constructing such a benchmark is challenging, as manual annotation is prohibitively expensive and prone to inconsistencies~\citep{loison2026vidore}. To this end, we developed a highly scalable, automated annotation pipeline. By synergizing advanced document parsing models with powerful MLLMs, this flexible pipeline ensures fine-grained precision and consistency, effectively laying the foundation for large-scale citation data generation while mitigating subjective human biases during the annotation process.

For evaluation, we move beyond answer accuracy and introduce a suite of Traceability Metrics. At its core is Strict Attributed Accuracy (\textbf{SAA}), a rigorous audit requiring the model to be correct in both its textual response and its visual evidence attribution. This ensures models are only rewarded when their answers are fundamentally grounded in correct evidence. For further diagnosis, we utilize \textbf{Recall} to evaluate evidence coverage and \textbf{Relevance} to verify logical alignment. 

Extensive experiments on 20 mainstream MLLMs reveal a pervasive and concerning phenomenon: \textbf{Attribution Hallucination.} As shown in Figure~\ref{fig:citevqa_example}c and Table~\ref{tab:main_results}, even top-tier models exhibit "pseudo-faithful" behavior, providing correct textual answers while citing entirely wrong locations. The SAA of state-of-the-art models like Gemini-3.1-Pro-Preview caps at 76.0, while leading open-source MLLMs fail to surpass the 25.0 threshold. This uncovers a severe logical fracture in current systems, further amplifying the risk of untraceable hallucinations, which must be resolved before deploying these models in critical real-world applications.

\paragraph{Contributions} Our main contributions are threefold:
\begin{itemize}
    \item \textbf{The CiteVQA Benchmark and Traceability Metrics}: We introduce an evaluation framework that transitions Doc-VQA from answer-only scoring to joint evidence-answer verification. Anchored by the Strict Attributed Accuracy (SAA) metric, we establish a rigorous standard for measuring element-level citation fidelity.
    \item \textbf{Scalable High-Fidelity Dataset Construction}: We design an automated data generation pipeline that resolves the cost and consistency bottlenecks of granular visual annotation. This approach enables the scalable creation of a robust dataset comprising 1,897 complex queries across 711 multi-page, multi-domain PDFs.
    \item \textbf{Discovery of the "Attribution Hallucination" Phenomenon}: Through a comprehensive audit of 20 leading MLLMs, we expose a critical vulnerability: models frequently output correct text while grounding it in entirely incorrect visual evidence. By demonstrating that state-of-the-art models cap at 76.0 SAA and leading open-source models fail to reach 25.0, we provide the critical instrumentation to advance trustworthy document intelligence.
\end{itemize}

\section{Related Work}

\paragraph{Document Visual Question Answering}
Document Visual Question Answering (Doc-VQA) has evolved from single-page comprehension~\citep{mathew2021docvqa, mathew2022infographicvqa, mishra2019ocr} to multi-page, full-document reasoning~\citep{tito2023hierarchical, ma2024mmlongbench, tanaka2023slidevqa}. Yet these benchmarks remain answer-centric, with evidence annotations largely restricted to the page level. Recent efforts have begun to attach bounding-box (BBox) evidence, but with clear limitations: ViDoRe~V3~\citep{loison2026vidore} relies on freely drawn boxes whose granularity varies across annotators (inter-annotator IoU of only 0.50), while SciEGQA~\citep{yu2026sciegqadatasetscientificevidencegrounded} is confined to 80 scientific papers, where its costly manual cross-annotation protocol does not scale. Meanwhile, domain-specific tasks like ChartQA~\citep{masry2022chartqa} and Charxiv~\citep{wang2024charxiv} target individual element types rather than diverse, layout-heavy real-world documents. CiteVQA instead provides element-level BBox citations with parser-standardized granularity over multi-page, multi-domain PDFs, jointly measuring answer accuracy and structural traceability; its fully automated pipeline scales readily to new domains without manual annotation.

\paragraph{Evidence Attribution and Citation Evaluation}
As hallucination remains a persistent threat in LLMs~\citep{wang2025rare, zhao2026retrieval}, attributing generated content to verifiable evidence has been extensively studied in text-only settings~\citep{gao2023enabling, lu2025med, chen2025gaps, zhang2025citalaw}. Multimodal attribution, however, remains largely confined to retrieval-augmented settings: MAVIS~\citep{song2026mavis} annotates long-form answers with sentence-level citations that point to retrieved documents, offering no spatial localization within them; VISA~\citep{ma2025visa} attaches bounding-box citations, yet only over single-page or stitched-image document representations, with attribution quality examined through a qualitative 50-case study; FinRAGBench-V~\citep{zhao-etal-2025-finragbench} introduces block-level visual citation within a retrieve-then-generate pipeline over a 60k-page financial corpus, but provides no ground-truth boxes for its QA set and relies on an evaluator MLLM to judge citation quality, itself validated only on a small annotated subset. These explorations differ substantially from real-world document intelligence, where users hold complete, layout-rich multi-page PDFs, and none systematically measures element-level evidence attribution---the very gap CiteVQA is built to close.

\paragraph{Document Intelligence Systems}
Early document intelligence systems predominantly adopted a coarse "page-level retrieval" paradigm. Systems like Colpali~\citep{faysse2024colpali}, VisRAG~\citep{yu2024visrag}, VDocRAG~\citep{tanaka2025vdocrag}, and M3DocRAG~\citep{cho2024m3docrag} segment documents into page-wise chunks, utilizing multimodal vector search for matching or localization. This macroscopic approach, however, falters on complex queries that demand precise, element-level grounding. Bolstered by the advanced reasoning capabilities of modern MLLMs~\citep{zheng2025deepeyes, zhang2025thyme}, recent architectures have transcended basic vector matching. SimpleDoc~\citep{jain2025simpledoc} refines precision through an iterative, summary-driven retrieval workflow, while agentic frameworks like DocLens~\citep{zhu2025doclens}, DocDancer~\citep{zhang2026docdancer}, and AgenticOCR~\citep{wang2026agenticocr} leverage tool-use to navigate from global pages down to localized visual elements. Yet, despite this systemic evolution toward fine-grained evidence extraction, evaluation paradigms have lagged. Existing benchmarks still primarily focus on end-answer accuracy, completely lacking the rigorous instrumentation needed to verify reasoning paths and visual traceability.

A more comprehensive discussion of related work, including text-only attribution literature and a systematic comparison with representative benchmarks, is provided in Appendix~\ref{Appendix: Extended Related Work}.

% Early document intelligence systems predominantly relied on the "page-level retrieval" paradigm, as exemplified by Colpali~\citep{faysse2024colpali}, VisRAG~\citep{yu2024visrag}, VDocRAG~\citep{tanaka2025vdocrag}, and M3DocRAG~\citep{cho2024m3docrag}. These works typically segment long documents into page-wise chunks and leverage multimodal vector matching for page localization. However, such coarse-grained approaches exhibit limited performance when handling complex queries that require precise element-level grounding. With the enhanced reasoning capabilities of MLLMs~\citep{zheng2025deepeyes, zhang2025thyme}, researchers have moved beyond simple vector matching: SimpleDoc~\citep{jain2025simpledoc} introduces an iterative retrieval workflow based on summary generation to refine matching precision, while recent efforts such as DocLens~\citep{zhu2025doclens}, DocDancer~\citep{zhang2026docdancer}, and AgenticOCR~\citep{wang2026agenticocr} push the boundaries further by integrating tool-use to achieve a transition from page navigation to fine-grained element-level evidence localization. Despite this systemic shift toward granular evidence extraction and utilization, existing Doc-VQA benchmarks still primarily focus on end-answer accuracy, generally lacking rigorous verification of model interpretability regarding reasoning paths and traceability of evidence association. 

\section{CiteVQA: A Benchmark for Faithful Evidence Attribution}
\label{CiteVQA}
\begin{figure}[tb]
\centering
\includegraphics[width=\textwidth]{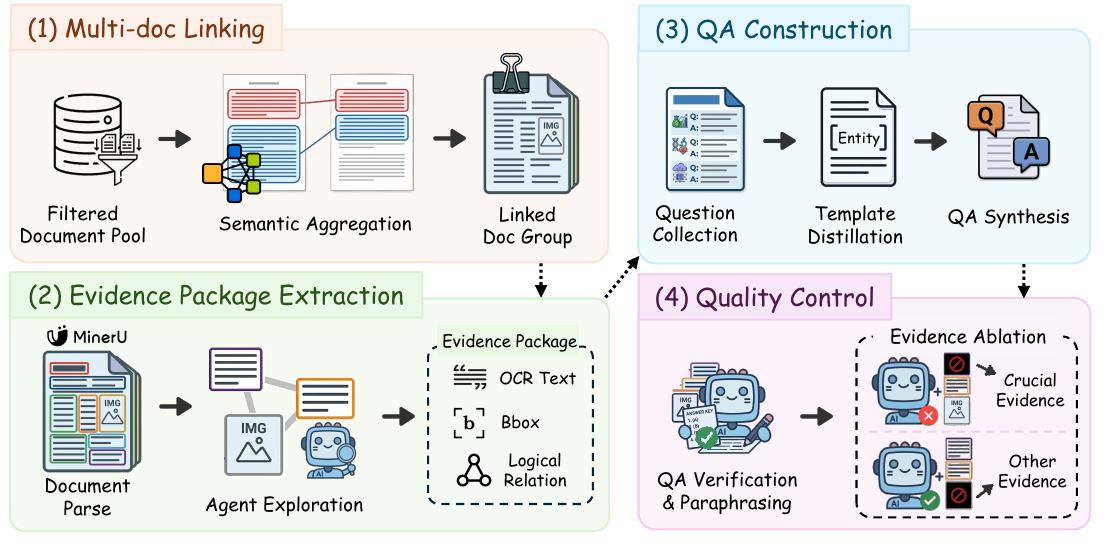}
\caption{The automated pipeline for CiteVQA. The workflow begins with Multi-doc Linking for semantic document aggregation, followed by Evidence Package Extraction, where intelligent agents navigate and link scattered MinerU parsing results into a cohesive evidence chain. To ensure authenticity, real-world QA pairs are distilled into templates to guide the automated synthesis of rigorous tasks. Finally, the pipeline implements MLLM-based verification and an evidence ablation procedure to precisely identify Crucial Evidence.}
\label{fig:citevqa_pipeline}
\end{figure}

To construct a high-quality benchmark with fine-grained evidence grounding, we develop an Automated Annotation Pipeline that streamlines the process from raw document parsing to complex question-citation generation. The overall workflow of this pipeline is illustrated in Figure~\ref{fig:citevqa_pipeline}. In the following subsections, we first provide a detailed introduction to each stage of the pipeline and then we present a comprehensive analysis of the Data Statistics to highlight the diversity and complexity of the CiteVQA benchmark.

\subsection{Document Collection}
To construct a highly representative and diverse evaluation benchmark, we designed a multi-stage automated filtering pipeline to systematically extract high-quality documents from a vast pool of heterogeneous data. Starting from a corpus of over 100 million raw PDF documents (primarily sourced from Common Crawl\footnote{\url{https://commoncrawl.org/}}; see Ethics Statement for compliance and ethical standards), we first pre-selected approximately 250k candidate documents through stratified sampling. These candidates then underwent a two-stage MLLM annotation scheme: (1) Coarse-grained stage, identifying the primary domain and language; and (2) Fine-grained stage, performing sub-category classification within each domain.

Ultimately, 711 documents were selected as the source for CiteVQA, achieving a balanced coverage across 7 domains and 30 sub-categories. This fully automated pipeline ensures both reproducibility and scalability.
\subsection{Question, Answer and Evidence Collection}
CiteVQA employs an end-to-end automated construction pipeline. The process first aggregates evidence through multi-document linking, then utilizes high-performance agents to extract complete evidence chains within fine-grained spatial contexts, and finally generates simulated real-world QA pairs through template-driven distillation.
\paragraph{Multi-Document Linking}
To overcome single-document limitations, we propose a linking strategy that aggregates cross-document evidence via semantic alignment. The system identifies candidates through vector similarity and utilizes an LLM to align section-level metadata, integrating isolated documents into logically connected groups $D$ (retaining single-document form if no associations exist). This provides a robust foundation for complex reasoning across multiple sources; see Appendix~\ref{Appendix: Details of Multi-Document Linking} for implementation.  
\paragraph{Evidence Package Extraction}
We utilize MinerU2.5~\citep{niu2025mineru2, wang2026mineru2} for deep document parsing to obtain fine-grained results containing document IDs, page numbers, bounding box (BBox) coordinates, and OCR content. Drawing inspiration from DocDancer~\citep{zhang2026docdancer} and WebSailor~\citep{li2025websailor}, we employ high-performance MLLMs (e.g., Gemini-3.0-Flash-Preview~\citep{team2023gemini}) as intelligent agents. These agents navigate the parsed BBox space to identify and concatenate supporting facts scattered across different pages or documents, ultimately aggregating them into a comprehensive Evidence Package.
\paragraph{QA Construction} 
To simulate real-world business scenarios effectively, we collect authentic questions from open-source datasets across various domains and distill them into a rich library of question templates---e.g., 138 templates for the financial domain alone (see Appendix~\ref{Appendix: Details of QA Construction}). During construction, high-performance MLLMs first select the most appropriate logical template based on the characteristics of the Evidence Package, subsequently synthesizing QA pairs automatically based on template constraints and core information within the evidence. This template-guided approach ensures both logical rigor and broad domain coverage.

\subsection{Quality Control and Assessment}
We implement a fully automated verification process to ensure dataset reliability. This includes Answerability Verification to confirm evidence sufficiency, Relevance Filtering to exclude common-knowledge questions, and an ablation-based procedure to identify "Crucial Evidence" for metric validity.  
\paragraph{Answerability Verification and Paraphrasing}
To eliminate invalid QA pairs, each candidate question undergoes a secondary confirmation by a powerful MLLM. Crucially, the verifier never sees the original document; it receives only the dependent evidence package: the bounding-box crops and their full-page screenshots, the OCR text, and the causal chain linking these elements to the answer. A QA pair is retained only if the model answers correctly from these inputs alone---this step verifies \emph{evidence sufficiency} rather than question difficulty. The retained questions are then paraphrased for linguistic diversity while strictly preserving the original intent.

\paragraph{Relevance Filtering and Crucial Evidence Identification}
To ensure the challenging nature of the dataset, we execute a "zero-document self-test" using Qwen3-VL-235B-A22B-Instruct~\citep{bai2025qwen3}: questions that the model can answer without any document context (classified as common-knowledge-based) are discarded.

For the core evidence chain determination, we designed an ablation-based crucial evidence identification procedure: each BBox element in the Evidence Package is masked individually before being presented to a powerful MLLM. If the model fails to derive the correct answer after a mask is applied, that element is labeled as \textbf{``Crucial Evidence.''} This procedure first guarantees the necessity and reliability of every crucial evidence, and on this basis further ensures the scientific validity of the subsequent Recall evaluation. Samples for which no single element proves crucial are discarded, so every retained question has at least one crucial evidence.

\paragraph{Remark} While our pipeline is fully automated to ensure scalability, we conducted human expert evaluation and auxiliary training validation to further guarantee the rigorous quality of the CiteVQA benchmark. Detailed procedures and results of these reliability assessments are provided in Appendix~\ref{Appendix: Details of Expert Evaluation} and~\ref{Appendix: Auxiliary Training Validation}.

\begin{table}[htbp] % 使用标准 table 环境，[htbp] 表示尽量放在当前位置
    \centering
    % 左侧：表格内容
    \begin{minipage}{0.48\textwidth}
        \centering
        \footnotesize
        \caption{Dataset Statistics}
        \label{tab:Dataset Statistics}
        \begin{tabular}{lc}
            \toprule
            \textbf{Statistic} & \textbf{Number} \\ 
            \midrule
            \textbf{Documents} & 711 \\
            \quad - Type (Macro/Micro) & 7 / 30 \\ 
            \quad - Avg./Median pages & 40.6 / 30.0 \\ 
            \quad - Language (EN/ZH) & 451 / 260 \\
            \midrule
            \textbf{Total questions} & 1,897 \\
            \quad - Single-doc & 987 (52.0\%) \\
            \quad - Multi (1-Gold) & 487 (25.7\%) \\
            \quad - Multi (N-Gold) & 423 (22.3\%) \\
            \midrule
            (Question Type) \\
            \quad - Complex Synthesis & 839 (44.23\%) \\
            \quad - Factual Retrieval & 499 (26.30\%) \\
            \quad - Multimodal Parsing & 352 (18.56\%) \\
            \quad - Quantitative Reasoning & 207 (10.91\%) \\
            \midrule
            (Crucial Evidence Source) \\
            \quad - Text & 2082 (70.12\%) \\ 
            \quad - Table & 653 (21.99\%) \\ 
            \quad - Image & 209 (7.04\%) \\ 
            \quad - Equation & 25 (0.84\%) \\ 
            \midrule
            Avg./Max. question length & 137.64 / 500 \\ 
            Avg./Max. answer length & 180.48 / 2976 \\ 
            Avg./Max. evidences & 2.57 / 10 \\
            \bottomrule
        \end{tabular}
    \end{minipage}
    \hfill % 撑开中间间距
    % 右侧：图片内容
    \begin{minipage}{0.45\textwidth}
        \centering
        \includegraphics[width=\linewidth]{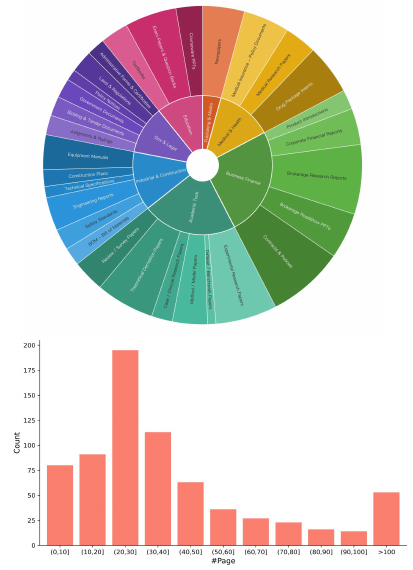} 
        \captionof{figure}{Distribution of documents. \textbf{Top}: Document type. \textbf{Bottom}: Page Number.}
        \label{fig:pdf_stats}
    \end{minipage}
\end{table}

\begin{figure}[htb]
\centering
\includegraphics[width=\textwidth]{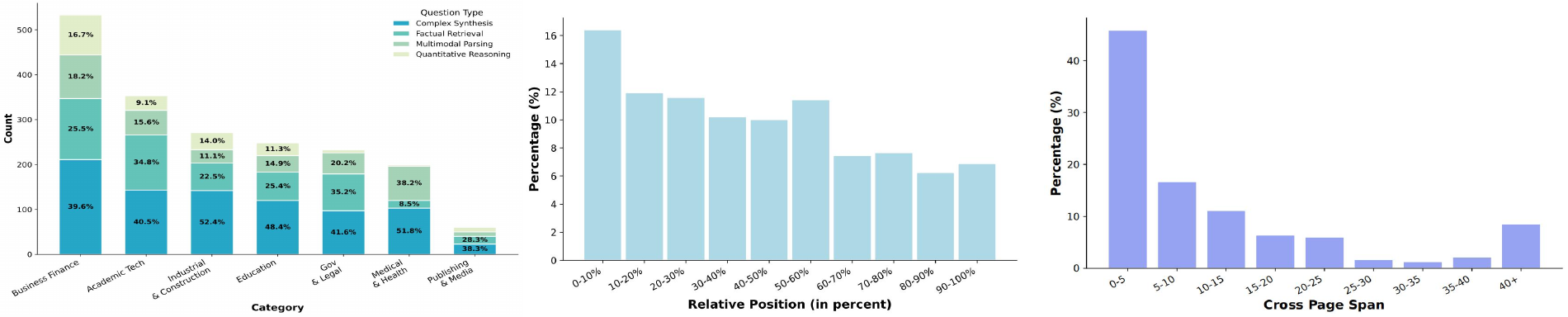}
\caption{Analysis of question types and evidence distribution in CiteVQA. \textbf{Left}: Domain-specific Question Types. \textbf{Middle}: Evidence Locality (in relative percent). \textbf{Right}: Cross-page Span, quantifying the number of pages spanned by evidences.}
\label{fig:question_statistics}
\end{figure}

\subsection{Dataset Overview and Analysis}
As summarized in Table~\ref{tab:Dataset Statistics} and Figures~\ref{fig:pdf_stats}-\ref{fig:question_statistics}, CiteVQA is a diverse benchmark comprising 711 documents across 7 macro-domains, with a realistic average length of 40.6 pages. The 1,897 questions cover varied scenarios including single-doc (52.0\%), multi-doc with one gold document (25.7\%), and multi-doc with multiple gold documents (22.3\%), spanning reasoning types from Complex Synthesis to Multimodal Parsing. Each task is grounded in an evidence package of 2.57 elements on average, and nearly 30\% of the crucial evidence elements are non-textual (tables, images, or equations). Evidence is uniformly distributed across document positions and often spans multiple pages, demanding robust long-context aggregation.

\begin{table}[h]
\centering
\scriptsize 
\renewcommand{\arraystretch}{1.25} 
\setlength{\tabcolsep}{1.0pt} 
\caption{\textbf{Comprehensive Evaluation of CiteVQA across Different Document Scenarios.} All scores are normalized to a 100-point scale; specifically, Rel. and Ans. scores (originally 0--5) are multiplied by 20 to facilitate direct comparison with Rec. and SAA. For each metric, the best and second-best results are highlighted in \highlightblue{blue} and \highlightgreen{green}, respectively.}
\label{tab:main_results}

\begin{tabular}{@{}l cccc cccc cccc >{\bfseries}c>{\bfseries}c>{\bfseries}c>{\bfseries}c@{}}
\toprule
\multirow{2.5}{*}{\textbf{Model}} & \multicolumn{4}{c}{\textbf{Single-Doc}} & \multicolumn{4}{c}{\textbf{Multi (1-Gold)}} & \multicolumn{4}{c}{\textbf{Multi (N-Gold)}} & \multicolumn{4}{c}{\textbf{Overall}} \\
\cmidrule(lr){2-5} \cmidrule(lr){6-9} \cmidrule(lr){10-13} \cmidrule(l){14-17}
& Rec. & Rel. & Ans. & SAA & Rec. & Rel. & Ans. & SAA & Rec. & Rel. & Ans. & SAA & Rec. & Rel. & Ans. & SAA \\ \midrule

% 分类1：闭源
\multicolumn{17}{c}{\textit{Closed-source MLLMs}} \\ \midrule
Gemini-3.1-Pro-Preview&\highlightblue{68.9}&\highlightblue{82.6}&\highlightgreen{86.7}&\highlightblue{76.0}&\highlightblue{69.4}&\highlightblue{84.3}&\highlightgreen{88.0}&\highlightblue{79.7}&\highlightblue{55.3}&\highlightblue{85.3}&\highlightgreen{82.8}&\highlightblue{71.6}&\highlightblue{66.0}&\highlightblue{83.6}&\highlightgreen{86.1}&\highlightblue{76.0} \\
Gemini-3-Flash-Preview&\highlightgreen{49.5}&\highlightgreen{76.8}&85.3&\highlightgreen{69.3}&\highlightgreen{42.1}&\highlightgreen{72.3}&86.0&\highlightgreen{61.8}&\highlightgreen{39.5}&\highlightgreen{77.0}&81.0&\highlightgreen{60.5}&\highlightgreen{45.4}&\highlightgreen{75.7}&84.5&\highlightgreen{65.4} \\ 
Gemini-2.5-Pro&31.5&61.7&83.0&49.4&25.4&58.3&84.1&48.9&20.0&57.3&78.0&39.2&27.4&59.8&82.2&47.0\\
GPT-5.4&35.9&69.8&\highlightblue{87.6}&61.7&25.7&62.1&\highlightblue{88.3}&56.9&25.7&68.2&\highlightblue{84.3}&55.1&31.0&67.5&\highlightblue{87.1}&59.0\\ 
GPT-5.2 &20.9&54.9&71.4&32.6&16.5&63.2&72.4&38.8&13.9&53.0&70.5&30.5&18.2&56.6&71.5&33.7 \\
Qwen3.6-Plus&9.8&26.7&87.1&20.2&5.9&24.6&87.3&18.5&4.6&21.3&81.2&9.8&7.7&25.0&85.9&17.5\\
Seed2.0-Pro&35.8&60.8&82.9&51.9&18.1&44.9&82.6&33.5&21.5&51.2&76.0&36.2&28.5&54.9&81.3&44.1 \\
GLM-5V-Turbo &18.3&31.2&50.0&14.1&11.7&25.2&44.4&9.8&10.2&28.8&54.3&13.0&14.9&29.2&49.6&12.8\\
\midrule

% 分类2：开源大模型
\multicolumn{17}{c}{\textit{Open-source Large MLLMs}} \\ \midrule
Kimi-K2.5&8.2&27.7&74.6&21.3&3.5&25.2&74.7&18.9&4.8&26.6&72.9&14.2&6.2&26.8&74.3&19.1 \\ 
Gemma-4-31B &10.9&31.0&65.7&16.4&14.0&41.0&80.6&29.8&10.4&37.7&67.2&17.8&11.6&35.0&69.8&20.2\\ 
Qwen3.5-397B-A17B&6.8&23.6&80.0&17.7&4.0&27.3&71.9&22.2&3.8&23.8&73.5&15.2&5.4&24.6&76.5&18.3\\
Qwen3.5-122B-A10B&5.9&19.4&78.3&16.0&1.7&20.9&69.0&17.0&1.9&15.6&68.1&9.2&3.9&19.0&73.6&14.8 \\
Qwen3.5-27B &7.0&25.0&79.3&17.1&3.1&28.3&73.1&22.6&3.9&22.6&69.9&11.6&5.3&25.3&75.6&17.3\\
Qwen3-VL-235B-A22B&15.2&37.8&75.0&25.0&6.2&33.8&68.2&21.6&8.1&31.4&70.9&17.8&11.3&35.3&72.3&22.5 \\
Qwen3-VL-32B&8.0&31.3&75.3&19.3&2.8&29.5&67.6&16.2&7.9&29.8&70.7&14.0&6.6&30.5&72.3&17.3 \\ 
\midrule

% 分类3：开源小模型
\multicolumn{17}{c}{\textit{Open-source Small MLLMs}} \\ \midrule
Gemma-4-26B-A4B &2.2&15.4&45.6&4.8&4.2&21.4&53.7&9.7&3.5&19.8&48.9&5.5&3.0&17.9&48.4&6.2 \\ 
Qwen3.5-35B-A3B  &2.7&12.6&82.3&9.2&0.5&17.3&73.9&15.6&0.6&12.1&65.5&8.3&1.7&13.7&76.4&10.7 \\
Qwen3.5-9B&2.5&11.7&73.2&8.1&0.3&20.7&58.4&17.7&0.8&14.6&53.5&10.7&1.6&14.7&65.0&11.1 \\
Qwen3-VL-30B-A3B  &5.6&15.4&65.4&8.9&0.9&15.9&54.4&8.2&1.7&11.0&63.5&6.4&3.5&14.6&62.2&8.2\\
Qwen3-VL-8B &1.8&17.6&67.0&8.8&0.0&13.6&53.3&6.8&0.3&9.3&56.4&5.2&1.0&14.7&61.2&7.5   \\

\bottomrule
\end{tabular}
\end{table}

\section{Evaluation}
\label{Evaluation}
\subsection{Evaluation Metrics}
\label{Evaluation Metrics}

To evaluate evidence attribution, we introduce a novel set of metrics assessing both answer correctness and trustworthiness in grounding predictions on verifiable evidence.

Formally, each sample is represented as $(D, Q, A_{\text{gt}}, \mathcal{B}_{\text{gt}})$, where $\mathcal{B}_{\text{gt}}$ is the set of ground-truth bounding boxes, further categorized into crucial ($\mathcal{B}_{\text{crucial}}$) and supplemental ($\mathcal{B}_{\text{other}}$) evidence. Each bounding box $b \in \mathcal{B}$ is defined by $(\text{doc\_idx}, \text{page\_idx}, x_1, y_1, x_2, y_2)$. The model output is $\hat{Y} = \{(A_1, b_1), \dots, (A_n, b_n)\}$, where $\mathcal{B}_{\text{pred}} = \{b_1, \dots, b_n\}$ denotes the predicted evidence set.

We define the following key metrics:

\textbf{Recall (Rec.)} Measures coarse-grained localization ability, computed at IoU@0.5 between predicted and crucial evidence: $\text{Rec.} = \frac{1}{|\mathcal{B}_{\text{crucial}}|} \sum_{b_{\text{gt}} \in \mathcal{B}_{\text{crucial}}} \mathbf{1}_{( \max_{b_{\text{pred}} \in \mathcal{B}_{\text{pred}}} \text{IoU}(b_{\text{pred}}, b_{\text{gt}}) \ge 0.5 )}$.

\textbf{Relevance (Rel.)} Measures how well the predicted evidence supports the answer, evaluated holistically by an LLM judge $\Jrel$ on a 0--5 scale: $\text{Rel.} = \Jrel(\{(A_1, b_1), \ldots, (A_n, b_n)\}, A_{\text{gt}}) \in [0, 5]$.

\textbf{Answer Correctness (Ans.)} Measures semantic matching between predicted and ground-truth answers via an LLM judge $\Jans$: $\text{Ans.} = \Jans(\{A_1, A_2, \dots, A_n\}, A_{\text{gt}}) \in [0, 5]$.

\textbf{Strict Attributed Accuracy (SAA)} A sample-level binary metric requiring both high-quality grounding and answer correctness: $\text{SAA} = \mathbf{1}_{(\text{Ans.} \ge 4 \land (\text{Rel.} \ge 4 \lor \text{Rec.} \ge 0.6))}$. Here the Rec.\ path enforces objective, IoU-based grounding, while the Rel.\ path acts as a safety net, crediting any citation that genuinely supports the answer even if it falls outside the crucial evidence.

In addition to the aforementioned metrics, we also evaluate $\text{Page}_{recall}$, Precision, and F1-score for a more comprehensive assessment of document localization. Owing to space limitations, their formal definitions and detailed evaluation results are deferred to the Appendix~\ref{Appendix: More Evaluation Metrics} and~\ref{Appendix: More Results of Experiments} .

\subsection{Experimental Setup}
\label{Experimental Setup}
We evaluated 20 state-of-the-art MLLMs, encompassing both leading proprietary and open-source models, on the CiteVQA benchmark. For input processing, models received sequential page screenshots via native APIs or OpenAI-compatible interfaces, with image resolutions adapted to their respective context window capacities (see Appendix~\ref{Appendix: Details of Experiments} for technical specifics). All models were tested using a unified prompt template with a sampling temperature of 1.0. For automated evaluation, we employed Qwen3-VL-235B-A22B as the primary judge (See Analysis of Judges in Appendix~\ref{Appendix: Analysis of Different Judges}).

\subsection{Main Results}

Table~\ref{tab:main_results} presents a comprehensive evaluation of state-of-the-art MLLMs on CiteVQA. Our analysis reveals several critical insights into the current state of faithful evidence attribution.

\paragraph{The "Attribution Hallucination" Phenomenon} A pervasive gap exists between answer accuracy (Ans.) and Strict Attributed Accuracy (SAA) across all tested models. Notably, while GPT-5.4 and Gemini-3-Flash achieve high answer scores (87.1 and 84.5), their SAA scores drop significantly to 59.0 and 65.4, respectively. This discrepancy confirms an "Attribution Hallucination" effect: models can extract information for a correct answer yet fail to link it to its specific spatial source. Our failure decomposition (Figure~\ref{fig:error_decomposition}) further shows that these failures are dominated by wrong-page and wrong-element citations rather than imprecise boxes.

\paragraph{Performance Disparity across Model Tiers} There is a stark performance hierarchy among different model categories. Closed-source MLLMs dominate the benchmark, with Gemini-3.1-Pro-Preview leading at an Overall SAA of 76.0. While GPT-5.4 excels in semantic answer correctness (87.1), it is surpassed by Gemini models in SAA, suggesting Gemini may have more robust native citation-alignment. In contrast, a significant "cliff" exists for Open-source Models, where the strongest (Qwen3-VL-235B) achieves an SAA of only 22.5. Small-scale MLLMs (e.g., Qwen3-VL-8B) struggle the most, with SAA scores often falling below 10.0. This underscores that deploying such small models in high-stakes domains—such as finance, law, or medicine—remains extremely risky, as they lack the fundamental grounding reliability required for professional auditing.  

\paragraph{Impact of Document Scenarios} Task difficulty scales with document complexity. While answer accuracy remains relatively stable across scenarios, attribution becomes markedly harder in multi-document settings. For example, Gemini-3.1-Pro’s Recall drops from 68.9 in Single-Doc tasks to 55.3 in Multi (N-Gold) scenarios. This multi-gold setting consistently yields the lowest SAA scores across the board, highlighting that cross-document evidence linking and complex spatial navigation remain significant frontiers for even the most advanced MLLMs.

\section{Analysis \& Discussion}
\subsection{Anatomy of SAA Failures}
\begin{wrapfigure}[11]{r}{0.45\textwidth}
    \centering
    \includegraphics[width=0.45\textwidth]{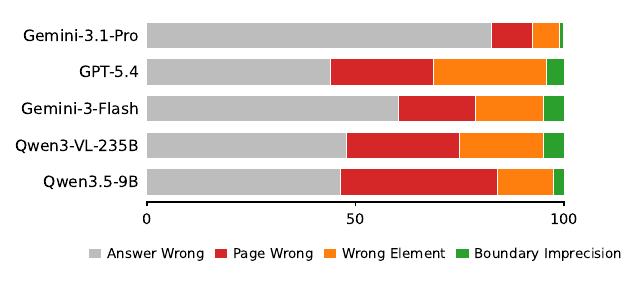}
    \caption{Decomposition of SAA$=0$ failures.}
    \label{fig:error_decomposition}
\end{wrapfigure}

\paragraph{What Drives SAA Failures?} A natural concern is whether SAA merely measures bounding-box drawing precision rather than evidence-finding ability. To answer this, we decompose every SAA$=0$ sample into four mutually exclusive categories (definitions in Appendix~\ref{Appendix: Error Decomposition}): (i)~\emph{Answer Wrong}, (ii)~\emph{Page Wrong}, (iii)~\emph{Wrong Element}, and (iv)~\emph{Boundary Imprecision}---the only one attributable to box-drawing. As shown in Figure~\ref{fig:error_decomposition}, category (iv) accounts for merely 1.0\%--5.0\% of failures across models, whereas wrong-page and wrong-element citations jointly constitute 34.8\%--51.7\%. Among answer-correct failures, the dominant mode is thus genuinely citing the wrong evidence, i.e., Attribution Hallucination, rather than imprecise box drawing.

\subsection{Further Analysis of Evidence Attribution}

Beyond the initial identification of attribution fallacies, we seek to further explore the nuanced relationship between Attribution and Accuracy.

\begin{table}[htbp]
    \centering
    \scriptsize % 统一缩小字号以适应并排布局
    
    % 左侧：趋势图
    \begin{minipage}[c]{0.48\textwidth}
        \centering
        \includegraphics[width=\textwidth]{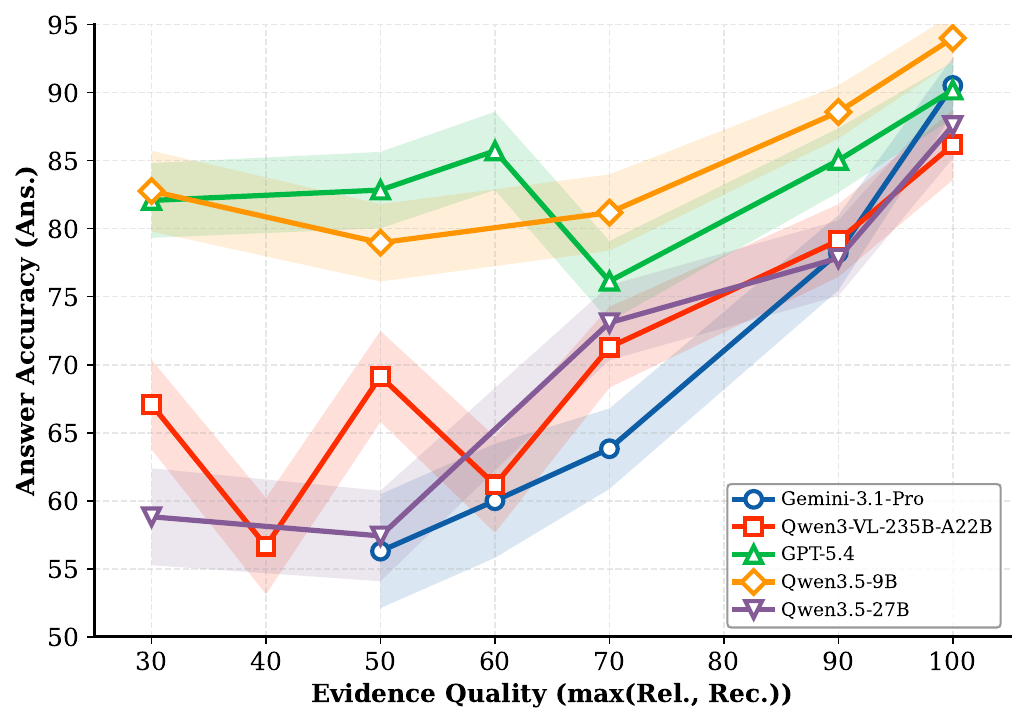}
        \captionof{figure}{Correlation between Evidence Quality and Answer Accuracy. The accuracy of MLLMs exhibits a fluctuating upward trend as evidence quality improves.}
        \label{fig:quality_vs_accuracy_100}
    \end{minipage}
    \hfill % 在两个 minipage 之间加入弹性间距
    % 右侧：消融实验表
    \begin{minipage}[c]{0.48\textwidth}
        \centering
        \caption{Ablation studies on Ground Truth (GT) pages and Gold documents, reporting answer correctness (Ans.). Narrowing the search space consistently leads to performance gains.}
        \label{tab:Ablation_Studies}
        \begin{tabular}{lcc}
            \toprule
            \textbf{Model} & \textbf{Base Setting} & \textbf{GT/Gold Setting} \\
            \midrule
            \rowcolor{gray!10} \multicolumn{3}{l}{\textit{Single-Doc vs. GT Pages}} \\
            Qwen3.5-27B    & 79.3  & 84.6 (\textcolor{teal}{+5.3}) \\
            Qwen3-VL-32B   & 75.3  & 79.9 (\textcolor{teal}{+4.6}) \\
            Qwen3.5-9B     & 73.2  & 75.2 (\textcolor{teal}{+2.0}) \\
            Qwen3-VL-8B    & 67.0 & 71.1 (\textcolor{teal}{+4.1}) \\
            \midrule
            \rowcolor{gray!10} \multicolumn{3}{l}{\textit{Multi (1-Gold) vs. 1 Gold Doc}} \\
            Qwen3.5-27B    & 73.1  & 81.6 (\textcolor{teal}{+8.5}) \\
            Qwen3-VL-32B   & 67.6  & 72.6 (\textcolor{teal}{+5.0}) \\
            Qwen3.5-9B     & 58.4  & 68.1 (\textcolor{teal}{+9.7}) \\
            Qwen3-VL-8B    & 53.3 & 66.7 (\textcolor{teal}{+13.4}) \\
            \bottomrule
        \end{tabular}
    \end{minipage}
\end{table}

\paragraph{Synergy between Attribution and Accuracy} Beyond serving as a metric for trustworthiness, faithful attribution appears to be positively correlated with the model’s reasoning success. As illustrated in Figure~\ref{fig:quality_vs_accuracy_100}, after bypassing the "Attribution Hallucination" zone (0-30 points), the Answer Accuracy (Ans.) tends to scale with Evidence Quality ($\max(Rel., Rec.)$). This upward trend provides an empirical hint that precise evidence localization might be more than just a post-hoc justification; it potentially acts as a functional foundation that facilitates correct answering in complex document-based tasks.
\begin{wrapfigure}[14]{r}{0.35\textwidth}
    \centering
    \includegraphics[width=0.35\textwidth]{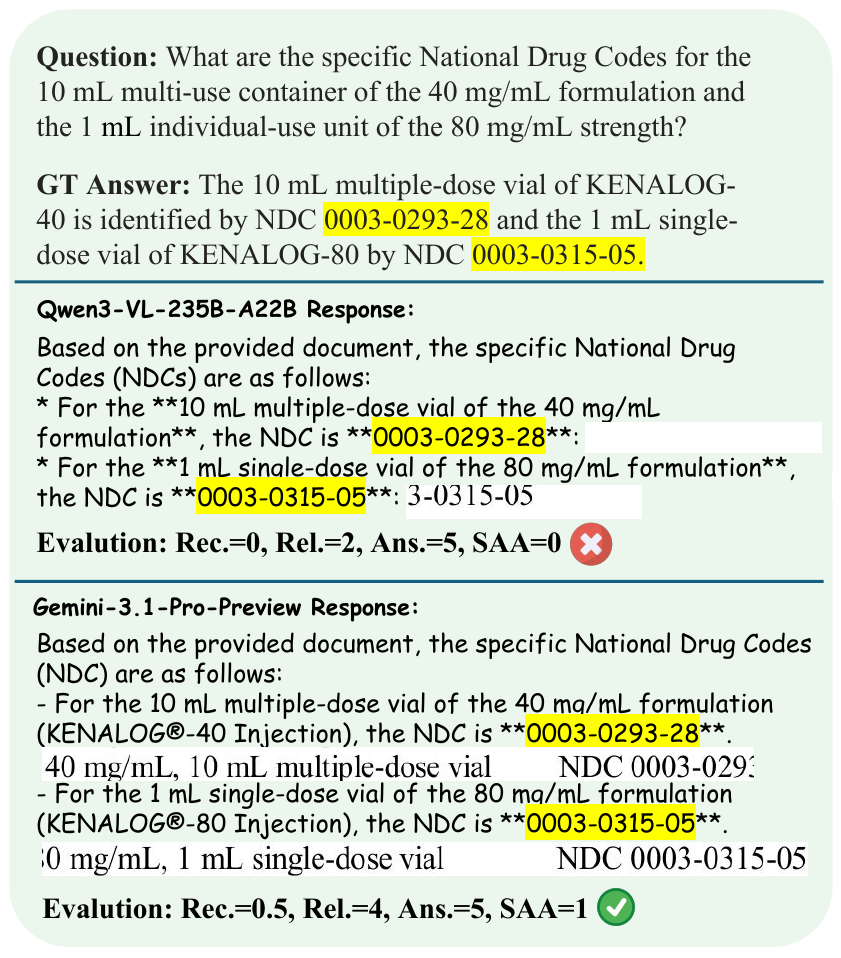}
    \captionof{figure}{A Typical Example.}
    \label{fig:Simple_Case}
\end{wrapfigure}

\paragraph{Evidence Attribution as a Potential Performance Driver} To further explore whether enhanced attribution could actively boost performance, we conducted ablation studies by narrowing the candidate search space (Table~\ref{tab:Ablation_Studies}). Restricting the context to GT-Pages consistently yielded gains (up to $+5.3\%$), while providing a single Gold Document in multi-document settings led to substantial leaps, such as $+13.4\%$ for Qwen3-VL-8B. These results offer preliminary evidence that some bottleneck in CiteVQA may also lie in the initial "localization" phase. It suggests a promising direction: if models can achieve higher autonomous evidence attribution recall, it might not only enhance transparency but also potentially unlock higher upper bounds for their answering capabilities.

\subsection{Case Study}
To intuitively illustrate the disparity between linguistic performance and attribution accuracy—specifically why some powerful MLLMs achieve high Ans. but low SAA scores—we provide a case study in Figure~\ref{fig:Simple_Case} (see Appendix~\ref{Appendix: Case Study} for more). For better visualization, cited coordinates are replaced with image crops.  

While Qwen3-VL-235B-A22B answers correctly (Ans.=5), it yields SAA=0 because its evidence crops are either blank or incomplete. In contrast, Gemini-3.1-Pro-Preview demonstrates faithful grounding; despite a slight offset in the first crop, its second is nearly perfect, resulting in SAA=1. This comparison underscores that a correct answer does not guarantee faithful evidence attribution.  

\section{Conclusion}
We introduced CiteVQA, a benchmark designed to advance trustworthy document intelligence by requiring models to provide element-level visual citations alongside answers. Leveraging an automated annotation pipeline, we constructed a large-scale dataset comprising 1,897 questions derived from 711 diverse PDFs. Our systematic audit of top-tier models reveals a critical "Attribution Hallucination" phenomenon, where correct answers are frequently paired with incorrect evidence. By exposing these hidden hallucinations, CiteVQA establishes a rigorous standard for developing interpretable and reliable multimodal systems in high-stakes domains.

\section*{AI use statement}
In this work, we used generative AI tools to generate the synthetic dataset and to support evaluation. Specifically, MLLMs (e.g., Gemini-3.0-Flash-Preview) are integral to the CiteVQA construction pipeline (Section~\ref{CiteVQA}): they act as agents for evidence package extraction, synthesize and paraphrase QA pairs from distilled templates, verify answerability, and perform the masking-ablation procedure for crucial evidence identification. During evaluation, an MLLM (Qwen3-VL-235B-A22B) serves as the judge for answer correctness and citation relevance (Section~\ref{Evaluation}); its agreement with human experts is analyzed in the appendix. Additionally, we used generative AI tools to edit the manuscript for readability and to assist with code implementation. All AI-assisted outputs were validated: constructed QA pairs and evidence underwent expert audit on a randomly sampled subset, and judge reliability was verified against human judgments. We have reviewed all AI-assisted work and take responsibility for the final content of this paper, including text, claims, and artifacts produced with the aid of generative AI.

\section*{Ethics statement}
\paragraph{Data Acquisition from Common Crawl} The 711 PDF documents included in the CiteVQA benchmark are sourced from Common Crawl, a neutral, non-profit public web archive. Our data acquisition process strictly adheres to the Common Crawl Terms of Use\footnote{\url{https://commoncrawl.org/terms-of-use}} and ensures all operations are conducted within the scope permitted by the Robots Exclusion Protocol, demonstrating our utmost respect for the intentions of original content distributors.

\paragraph{Adherence to Common Standards for Data Distribution}
Regarding data derived from Common Crawl, CiteVQA strictly follows the common consensus and academic norms of the multimodal and document intelligence fields. Drawing on the distribution paradigms of landmark works—such as T5~\citep{raffel2020exploring}, MMC4~\citep{zhu2023multimodal}, OBELICS~\citep{laurencon2023obelics}, and CCpdf~\citep{turski2023ccpdf}, which focuses specifically on PDF parsing—we have implemented an academically recognized compliance workflow (i.e., distributing only public download links).

We have open-sourced the structured metadata and spatial coordinates (Bounding Boxes) annotated by our automated data pipeline. We ensure that our PDF usage logic remains consistent with large-scale open-source datasets like LAION~\citep{schuhmann2021laion} to promote academic reproducibility while maintaining the compliance of the content distribution chain.

\paragraph{Copyright Respect and Protection of Rights Holders} All PDF documents in this dataset sourced from Common Crawl are clearly attributed in the repository's metadata. We hold the legal rights of original copyright holders in the highest regard. If any owner of the relevant documents believes that the indexing or usage within this benchmark is inappropriate, please contact us. We commit to cooperating promptly with the removal or updating of the relevant content upon verification of the rights holder's identity.

\paragraph{Vision: Advancing Transparent and Trustworthy AI Auditing} The core vision of CiteVQA is to address the issue of "evidence Attribution Hallucination" in MLLMs when processing complex documents. Inspired by the ethical guidelines of OBELICS~\citep{laurencon2023obelics}, we firmly believe that establishing transparent, traceable, and open-source benchmarks is key to building a responsible AI ecosystem. By constructing this evidence-chain benchmark, we aim to provide the community with an auditable and reproducible research tool, pushing global document intelligence research toward a more faithful and transparent future.

\section*{Reproducibility Statement}
We are committed to making CiteVQA reproducible. The complete benchmark dataset---all 1,897 questions with ground-truth answers and element-level bounding-box annotations---is provided in the Supplementary Material. Document metadata with public download links, the evaluation code, and the question-template library (e.g., the full 138-template financial set) are available at \url{https://github.com/opendatalab/CiteVQA}. The automated construction pipeline is described in detail in Section~\ref{CiteVQA} and Appendix~\ref{Appendix: Details of CiteVQA Pipeline}, with the key prompts provided in Appendix~\ref{Prompt Templates}, enabling independent re-implementation. The evaluation protocol, including per-model input configurations, metric definitions, and judge setup, is specified in Section~\ref{Evaluation} and Appendix~\ref{Details & More Results of Experiments}.

%\clearpage
%\newpage
{\small
\bibliography{paper}
\bibliographystyle{iclr2027_conference}
}

\clearpage
\newpage
\appendix
\section*{Appendix}
\label{sec:appendix}
\startcontents[sections]
\printcontents[sections]{l}{1}{\setcounter{tocdepth}{3}}

\newpage
\section{Extended Related Work and Benchmark Comparison}
\label{Appendix: Extended Related Work}
\begin{table}[h]
\centering
\small
\setlength{\tabcolsep}{5pt}
\renewcommand{\arraystretch}{1.15}
\caption{CiteVQA vs.\ representative Doc-VQA benchmarks. \textit{Gran.}: evidence granularity (\textbf{P} page-level, \textbf{B} human-drawn bounding boxes, \textbf{E} parser-determined element-level boxes). \textit{Joint}: answer and sub-page (bounding-box) citation scored by a single sample-level metric.}
\label{tab:benchmark_compare}
\begin{tabular}{@{}lcccc@{}}
\toprule
\textbf{Benchmark} & \textbf{\#Docs} & \textbf{Avg.\ Pg.} & \textbf{Gran.} & \textbf{Joint} \\
\midrule
DocVQA~\citep{mathew2021docvqa}             & 12{,}767 & 1.0  & P & $\times$ \\
InfoVQA~\citep{mathew2022infographicvqa}    & 5{,}485  & 1.0  & P & $\times$ \\
MP-DocVQA~\citep{tito2023hierarchical}      & 6{,}000  & 8.3  & P & $\times$ \\
MMLongBench-Doc~\citep{ma2024mmlongbench}   & 135      & 47.5 & P & $\times$ \\
SlideVQA~\citep{tanaka2023slidevqa}         & 2{,}619  & 20.0 & P & $\times$ \\
DUDE~\citep{Van_Landeghem_2023_ICCV}            & 5{,}019  & 5.7  & B & $\times$ \\
ViDoRe~V3~\citep{loison2026vidore}          & 190      & 137.0 & B & $\times$ \\
SciEGQA~\citep{yu2026sciegqadatasetscientificevidencegrounded} & 80 & 24.3 & B & $\times$ \\
\midrule
\textbf{CiteVQA (ours)}                     & \textbf{711} & \textbf{40.6} & \textbf{E} & \checkmark \\
\bottomrule
\end{tabular}
\end{table}

\paragraph{What Element-level Means and Why It Matters}
In CiteVQA, an \textbf{element} is a typed, semantically complete document unit---a full paragraph, a complete table, a complete figure, or a note---delimited by MinerU's deterministic parser: parsing the same document twice yields nearly identical boundaries, so the evidence ground truth is unambiguous and supports rigorous IoU-based evaluation. Being fully automated and domain-agnostic, this pipeline also transfers readily to new domains without redesigning annotation protocols. Human-drawn boxes enjoy neither property: DUDE~\citep{Van_Landeghem_2023_ICCV} similarly relies on annotator-drawn boxes, restricted to its extractive subset; ViDoRe~V3~\citep{loison2026vidore} reports that annotators ``differed in granularity,'' reaching an inter-annotator IoU of only 0.50, and SciEGQA~\citep{yu2026sciegqadatasetscientificevidencegrounded} must rely on a costly cross-annotation protocol (IoU $>0.9$ agreement with adjudication) that does not scale---its authors explicitly restrict automatically constructed data to training and use manual annotation for evaluation. SlideVQA~\citep{tanaka2023slidevqa} also ships bounding boxes, but they are generic layout annotations---OCR/layout-parser regions with category labels, used only as input features at training time---and its question-level evidence annotation consists solely of slide indices; no question is linked to any sub-page region, so bounding-box-level evidence evaluation is not supported.

\paragraph{Evidence Attribution in General Settings}
In text-only scenarios, ALCE~\citep{gao2023enabling} automatically evaluates whether citations in LLM-generated text genuinely support the claims, while domain-specific systems require answers to be traced to clinical guidelines or legal statutes~\citep{lu2025med, chen2025gaps, zhang2025citalaw}. In multimodal retrieval-augmented generation, MRAMG-bench~\citep{yu2025mramg} evaluates answers that interleave text with images drawn from the corpus, using a statistically grounded suite of metrics. Although these pioneering efforts do not involve citation within real PDF documents, they lay important foundations and provide valuable references for the design of CiteVQA.

\newpage
\section{Details of CiteVQA Pipeline}
\label{Appendix: Details of CiteVQA Pipeline}

\subsection{Details of Multi-Document Linking}
\label{Appendix: Details of Multi-Document Linking}

\paragraph{Semantic Profiling and Dense Retrieval}
To mitigate semantic truncation caused by directly embedding long documents, we first use MLLM to construct a semantic profile for each document $d_i$, extracting metadata such as document type, core thesis, and section units. These profiles serve as high-level descriptors that capture the global context of the document beyond simple text snippets. The extracted metadata is then mapped to normalized vectors via an encoder. For an anchor document $d_a$, the top-$K_{doc}$ (default 5) candidate documents are selected based on cosine similarity to form the candidate pool $C_a$, ensuring that only the most contextually relevant documents are considered for expensive fine-grained analysis. 

\paragraph{Fine-Grained Alignment via LLM}
While coarse retrieval identifies document-level relevance, cross-document QA requires precise segment-level evidence chains. We input all section units from both the anchor and candidate documents into an LLM to perform chain-of-thought (CoT) cross-document matching. The model is prompted to reason through the structural hierarchy of each document to find logical bridges between them. The model outputs structured association groups, including the anchor section, candidate section, a similarity score $s \in [0,1]$, and a brief rationale for the connection. We limit the output to a maximum of 5 matching groups with 1–3 related segments per group to maintain high information density. The system retains the best matches in descending order of scores and returns an empty list when no reliable match is found, thereby filtering out noisy or coincidental associations. 

\paragraph{Spatial Mapping and Evidence Synthesis}
The matched pages from diverse sources are assembled into a synthetic document, which serves as the workspace for generating complex QA pairs spanning $\ge 2$ source documents. A critical component of this process is the element-level bijective function $f_{map}$, which maintains a persistent link between the synthetic layout and the original files. This function maps the synthetic evidence bounding boxes back to their original spatial coordinates in the source PDF. By ensuring that every byte of synthesized evidence is traceable to its original page, the system fundamentally eliminates visual hallucinations and ensures the absolute fidelity of the citation annotations.

\begin{table}[htbp]
\centering
\small
\caption{Distribution of Multi-domain Open-source Datasets}
\label{tab:domain_datasets}
\begin{tabular}{lll}
\toprule
\textbf{Domain} & \textbf{Dataset} & \textbf{License} \\
\midrule
Education & CK-12 TQA \citep{ck12_tqa_multimodal2025} & CC BY-NC 3.0 \\
Academic Tech & SPIQA \citep{pramanick2024spiqa} & CC BY 4.0 \\
\multirow{2}{*}{Medical \& Health} & MedQA \citep{jin2020disease} & MIT \\
 & PubMedQA \citep{jin2019pubmedqa} & MIT \\
Business Finance & ViDoRe V3 \citep{loison2026vidore} & CC BY 4.0 \\
\multirow{2}{*}{Industrial \& Construction} & MaintNorm \citep{bikaun2024maintnorm} & MIT \\
 & ViDoRe V3 \citep{loison2026vidore} & CC BY 4.0 \\
Gov \& Legal & PolicyBench \citep{foo2025know} & OpenRail \\
\bottomrule
\end{tabular}
\end{table}

\subsection{Details of Template Distillation}
\label{Appendix: Details of QA Construction}

To simulate real-world business scenarios effectively, we collected diverse problems from multi-domain open-source datasets and distilled them into a rich library of question templates. The template sets are substantially larger than the illustrative examples in Table~\ref{tab:template_examples}: the financial domain alone contains 138 templates---41 Factual Retrieval, 33 Multimodal Parsing, 33 Quantitative Reasoning, and 31 Complex Synthesis---and every domain has a template set of comparable scale. 

Table~\ref{tab:domain_datasets} details the distribution of the datasets used in our framework, covering 6 key domains to ensure broad representation. We would like to express our sincere gratitude to the contributors of these open-source projects for their invaluable support to the research community.

\begin{table}[htbp]
\centering
\small
\caption{Classification and Examples of Templates}
\label{tab:template_examples}
\begin{tabularx}{\textwidth}{l|X}
\toprule
\textbf{Category} & \textbf{Template and Representative Example} \\
\midrule
\textbf{Factual Retrieval} & \textbf{Template:} What are the reported [Revenue], [Cost], and [Profit] values for [Entity/Segment] in [Period] across the core statements? \\
 & \textbf{Example:} What are the reported revenue, cost, and operating profit values for the cloud segment in FY2024 across the disclosed statements? \\
\midrule
\textbf{Complex Synthesis} & \textbf{Template:} How do the [Mechanism Explanation], [Comparative Data], and [Forward Catalyst] jointly support the report's [Investment Conclusion]? \\
 & \textbf{Example:} How do the IRAK4 mechanism explanation, competitor comparison data, and upcoming readout catalysts jointly support the report's investment conclusion on the pipeline? \\
\midrule
\textbf{Quantitative Reasoning} & \textbf{Template:} Using the reconciliation table, derive [Adjusted Metric] from [GAAP Metric] by applying all listed adjustments. \\
 & \textbf{Example:} Using the reconciliation table, derive adjusted EBITDA from operating profit by applying all listed add-backs and deductions. \\
\midrule
\textbf{Multimodal Parsing} & \textbf{Template:} Extract and stitch the complete [Cross-page Table], preserving all continued rows and columns across pages. \\
 & \textbf{Example:} Extract and stitch the complete licensing-deals table across all continuation pages. \\
\bottomrule
\end{tabularx}
\end{table}

\subsection{Details of Expert Evaluation}
\label{Appendix: Details of Expert Evaluation}
Despite the automated production, we invited several PhD-level experts to conduct sampling audits of 200 randomly selected CiteVQA outputs, focusing on question difficulty, answer quality, and the quality of crucial evidence. To ensure consistency in the evaluation standard, the experts followed the same prompt templates as the models, the details of which are provided in Appendix~\ref{Prompts for CiteVQA Pipeline}. The audit results (Table~\ref{tab:Expert Evaluation}) confirm the high quality of the automated pipeline, demonstrating appropriate question difficulty and high-quality annotation.

\begin{table}[htbp]
    \centering
    \footnotesize
    \caption{Annotation evaluation results on 200 sampled CiteVQA instances (5-point Likert scale, averaged across human experts).}
    \label{tab:Expert Evaluation}
    \begin{tabular}{lccc}
        \toprule
        \textbf{Judge / Metric} & 
        \makecell{\textbf{Question}\\ \textbf{Difficulty}} & 
        \makecell{\textbf{Answer}\\ \textbf{Quality}} & 
        \makecell{\textbf{Evidence}\\ \textbf{Quality}} \\
        \midrule
        Gemini-3-Flash   & 2.81 & 4.57 & 4.93 \\
        Qwen3-VL-235B       & 2.73 & 4.62 & 4.89 \\
        Human Expert            & 2.97 & 4.43 & 4.91 \\
        \bottomrule
    \end{tabular}
\end{table}

\subsection{Details of Auxiliary Training Validation}
\label{Appendix: Auxiliary Training Validation}
To assist in validating the effectiveness of the automated pipeline in real-world training, we conducted an alignment experiment based on the ViDoRe V3~\citep{loison2026vidore} corpus in Table~\ref{tab:retrieval_results}. Following the same PDFs, we generated 3k samples via CiteVQA and compared their performance against the original 5k human-annotated samples in AgenticOCR~\citep{wang2026agenticocr}  SFT training. Overall, CiteVQA Pipeline nearly reaches the performance level of human-annotated data. 

\begin{table*}[htbp]
\centering
\footnotesize
\setstretch{1.2}
\caption{Performance comparison on FinRAGBench-V~\citep{zhao-etal-2025-finragbench} (subset with bounding boxes) and the held-out test set of ViDoRe V3. 
$\mathbf{Page_{acc}}$ measures page-level judgment ability; $\mathbf{Recall_{min}}$ indicates coarse-grained localization; 
$\mathbf{Recall_{EM}}$ reflects exact-grained localization. On FinRAGBench-V, CiteVQA Pipeline (3k) achieves comparable or 
slightly better performance than Vidore Original (5k). On ViDoRe V3, Vidore Original shows a slight advantage. 
}
\resizebox{\textwidth}{!}{
\begin{tabular}{lccccc|ccccc}
\toprule
\multirow{2}{*}{\textbf{Training Data}} 
& \multicolumn{5}{c|}{\textbf{FinRAGBench-V (subset w. bbox)}} 
& \multicolumn{5}{c}{\textbf{ViDoRe V3 (test set)}} \\
\cmidrule(lr){2-6} \cmidrule(lr){7-11}
& $\mathbf{Page_{acc}}$ & $\mathbf{Recall_{min}}$ & $\mathbf{Prec_{min}}$ & $\mathbf{F1_{min}}$ & $\mathbf{Recall_{EM}}$
& $\mathbf{Page_{acc}}$ & $\mathbf{Recall_{min}}$ & $\mathbf{Prec_{min}}$ & $\mathbf{F1_{min}}$ & $\mathbf{Recall_{EM}}$ \\
\midrule
Vidore Original (5k) & 97.7 & 83.0 & 85.2 & 82.7 & 35.4 & 94.7 & 83.4 & 82.2 & 81.0 & 48.3  \\
CiteVQA Pipeline (3k) & 97.7 & 82.8 & 86.0 & 81.3 & 40.6 & 93.5 & 79.5 & 82.4 & 78.8 & 45.3  \\
\bottomrule
\end{tabular}
}
\label{tab:retrieval_results}
\end{table*}

In the following, we describe the technical implementation of our distillation strategy, the specific tools provided to the model during training, and the rejection sampling criteria used to ensure the high quality of the resulting trajectories.

\paragraph{Trajectory Distillation via Rejection Sampling} We follow the same SFT training data distillation pipeline as AgenticOCR~\citep{wang2026agenticocr}. Specifically, we use the CiteVQA pipeline to generate a set of synthetic data from the original PDF files of ViDoRe V3. The data format is similar to that of the original ViDoRe dataset, namely (I, Q, A, E), representing Image, Question, Answer, and Evidence Bbox, respectively. 

For both batches of data, we adopt the AgenticOCR approach: we first equip the model with an image\_zoom\_and\_ocr\_tool (allowing the model to zoom into image regions and obtain OCR results), and then perform rejection sampling on the trajectories generated by Gemini-3-Pro-Preview based on an IoU threshold. This yields 3k and 5k high-quality samples, respectively.

\paragraph{Training Setup} We fine-tune \textit{Qwen3-VL-8B-Instruct}~\citep{bai2025qwen3} as the student model for both data batches. We follow the same training protocol as AgenticOCR: only the tokens generated by the assistant (including reasoning steps and tool calls) contribute to the loss; tokens corresponding to user prompts and tool observations are masked out. The hyperparameters are set as follows: a learning rate of \(1 \times 10^{-5}\), training for 6 epochs on 8 H200 GPUs.

\paragraph{Model Evaluation} We evaluate on two test sets. The first is the FinRAGBench-V~\citep{zhao-etal-2025-finragbench} subset with bounding box annotations (approximately 200 samples). The second is the held-out, manually annotated test set from ViDoRe V3 that is completely disjoint from the training set (approximately 400 samples). The evaluation metrics are identical to those used in AgenticOCR. The final results are reported in Table~\ref{tab:retrieval_results}.

\newpage
\section{Details \& More Results of Experiments}
\label{Details & More Results of Experiments}
\subsection{Details of Experimental Setup}
\label{Appendix: Details of Experiments}
\paragraph{Input Processing Details} For the Gemini series, we utilized the native File API. 

For other models, PDF documents were converted to 150 DPI screenshots. To ensure fairness across different context limits:

\begin{itemize}
    \item \textbf{Long-context Models:} Provided with original 150 DPI screenshots.
    \item \textbf{Standard-context Models:} Screenshots were adaptively downscaled according to the specific context constraints of each model family (details provided in Table \ref{tab:appendix_model_specs}).
\end{itemize}

\begin{table}[htbp]
\centering
\small
\caption{Model Categorization and Detailed Input Processing Strategies}
\label{tab:appendix_model_specs}
\begin{tabularx}{\textwidth}{l p{3.5cm} X}
\toprule
\textbf{Category} & \textbf{Models} & \textbf{Resolution and Processing Strategy} \\
\midrule
\textbf{Gemini Series} & Gemini-3.1-Pro-Preview, \newline Gemini-3-Flash-Preview, \newline Gemini-2.5-Pro~\citep{team2023gemini} & \textbf{Native File API:} Directly processed via the Google Cloud document interface without manual rasterization. \\
\midrule
\textbf{1M Context} & GPT-5.4, GPT-5.2~\citep{achiam2023gpt}, \newline Qwen3.6-Plus & \textbf{Full Resolution:} 150 DPI page screenshots provided as-is to leverage the expansive context window. \\
\midrule
\textbf{256k Context} & Qwen3.5 Family, \newline Qwen3VL Family~\citep{bai2025qwen3}, \newline Gemma4 Family~\citep{team2024gemma}, \newline Kimi-K2.5~\citep{team2024gemma}, \newline Seed-2.0-Pro~\citep{seed2026seed18modelcardgeneralized}  & \textbf{Standard Scaling:} Screenshots are adaptively downscaled to a maximum of $1024 \times 1024$ pixels, preserving the original aspect ratio. \\
\midrule
\textbf{200k Context} & *Only for GLM-5V-Turbo~\citep{zeng2026glm} & \textbf{Compact Scaling:} Screenshots are adaptively downscaled to a maximum of $768 \times 768$ pixels to prevent context overflow while maintaining structural integrity. \\
\bottomrule
\end{tabularx}
\end{table}

\begin{table}[htbp]
\centering
\small
\caption{Impact of input resolution on CiteVQA performance using Qwen3-VL-235B-A22B. We compare our standard scaling ($1024^2$) against reduced pixel budgets to evaluate the sensitivity of evidence attribution to visual clarity. SAA highlights the precipitous drop in grounding reliability as resolution decreases.}
\label{tab:resolution_impact}
\begin{tabular}{@{}l c cccc@{}}
\toprule
\textbf{Resolution Strategy} & \textbf{Total Pixels} & \textbf{Rec.} & \textbf{Rel.} & \textbf{Ans.} & \textbf{SAA} \\ \midrule
Full Resolution (Standard) & $1024^2$ ($1.0\times$) &11.3&35.3&72.3&22.5  \\
Half-Pixel Scaling & $1024^2 / 2$ ($\approx 724^2$) &4.2&23.6&66.8&11.8 \\
Quarter-Pixel Scaling & $1024^2 / 4$ ($512^2$) &1.6&17.2&53.5&5.3 \\ \bottomrule
\end{tabular}
\end{table}

\paragraph{Trade-off Analysis of Input Resolution}
The results in Table \ref{tab:resolution_impact} justify our choice of $1024 \times 1024$ as the standard resolution for evaluation. We observe that while the answer accuracy (Ans.) decreases moderately with lower resolutions, the evidence attribution metrics—particularly Rec. and SAA—exhibit a sharp, non-linear collapse. For instance, halving the total pixels (from $1024^2$ to $\approx 724^2$) leads to a near-50\% reduction in SAA (from 22.5\% to 11.8\%), indicating that precise localization is highly sensitive to visual fidelity. Although further increasing the resolution might yield marginal gains, $1024 \times 1024$ represents a critical "saturation point" for most current MLLMs. Exceeding this threshold would surpass the native token limits and internal position embedding constraints of many models (e.g., the Qwen3VL and Gemma families). Thus, our standard scaling maintains an optimal balance between preserving fine-grained document details and adhering to the architectural limits of diverse model families.
Finally, these resolution constraints act unevenly across model families: proprietary systems ingest natively rendered or full-resolution pages, whereas open-source models are limited by their context windows to downsampled inputs, so part of the performance gap between model tiers (Section~\ref{Evaluation}) may reflect input fidelity rather than capability alone.

\paragraph{Inference Settings} All experiments were conducted using a unified prompt (See Appendix~\ref{Prompts for CiteVQA Evaluation}). The maximum output length was capped at 4,096 tokens to allow for extensive reasoning. For the Qwen3VL family, the "Instruct" versions were consistently used. For the Qwen3.5 series, the Thinking mode was enabled by default. All GPT models were configured with the highest reasoning effort (xhigh), and Gemini models were run with the maximum thinking mode setting.

\paragraph{Deployment of Open-source Models}We utilized a standardized inference infrastructure consisting of 8$\times$NVIDIA H200 GPUs to ensure consistent latency and sufficient VRAM for high-resolution document processing.

\subsection{Analysis of Different Judges}

\paragraph{Validation of Automated Evaluation via Human Study}
To verify the reliability of our automated evaluation pipeline, we conducted a human expert study on 200 randomly selected samples, comparing human scores against those generated by Gemini-3-Flash-Preview and Qwen3-VL-235B-A22B. As detailed in Table~\ref{tab:model-eval}, we applied the Friedman test—a non-parametric statistical test—to determine if any significant differences existed between the judges. The resulting $p$-values for both Rel. and Ans. consistently exceeded the 0.05 threshold (ranging from 0.14 to 0.53) across different inference models. These results indicate no statistically significant deviation between automated judges and human experts; moreover, the absolute differences in mean scores are minimal (0.02--0.16 on a 5-point scale across both metrics and both judges), providing direct numerical evidence that LLM-based scoring closely tracks human judgment. 
\label{Appendix: Analysis of Different Judges}
\begin{table}[htbp]
\centering
\footnotesize
\caption{Comparison between automated LLM judges and human expert scores. $p$-values $> 0.05$ (Friedman test) indicate no statistically significant difference between automated judges and human experts.}
\label{tab:model-eval}
\begin{tabular}{l w{c}{0.8cm} w{c}{0.8cm} w{c}{0.8cm} w{c}{0.8cm}}
\toprule
\multirow{2}{*}{Judge / Infer Model} & \multicolumn{2}{c}{GPT-5.4} & \multicolumn{2}{c}{Gemini-3.1-Pro} \\
\cmidrule(lr){2-3} \cmidrule(lr){4-5}
 & Rel. & Ans. & Rel. & Ans. \\
\midrule
Gemini-3-Flash-Preview & 2.87 & 4.51 & 4.06 & 4.67 \\
\addlinespace
Qwen3-VL-235B-A22B       & 3.08 & 4.42 & 4.12 & 4.57 \\
\addlinespace
Human Expert             & 2.92 & 4.44 & 4.03 &  4.59\\
\midrule
P-value (Friedman Test) & 0.16 & 0.14 & 0.53 &  0.21 \\
\bottomrule
\end{tabular}
\end{table}

\subsection{More Evaluation Metrics}
\label{Appendix: More Evaluation Metrics}
To provide a multi-dimensional perspective on document localization and evidence attribution, we introduce several supplementary metrics. These indicators offer a more granular analysis of the model's performance in identifying relevant document components.

\textbf{Page-level Recall (Page. / $\text{Page}_{recall}$)} This metric assesses the model's coarse-grained ability to locate the correct pages containing the necessary evidence. A predicted evidence is considered a "page hit" if its page index matches any page index in the set of crucial evidence.
\[ 
\text{Page.} = \frac{|\{p \in \mathcal{P}_{\text{crucial}} \mid \exists \hat{p} \in \mathcal{P}_{\text{pred}}, \hat{p} = p\}|}{|\mathcal{P}_{\text{crucial}}|} 
\]
where $\mathcal{P}_{\text{pred}}$ and $\mathcal{P}_{\text{crucial}}$ denote the sets of page indices from the predicted and ground-truth crucial evidence, respectively.

\textbf{Precision (Prec.)} While Recall focuses on coverage, Precision measures the spatial accuracy of the predicted bounding boxes relative to the entire evidence set $\mathcal{B}_{\text{gt}}$ (including both crucial and auxiliary evidence). It penalizes the model for generating redundant or irrelevant boxes:
\[ 
\text{Prec} = \frac{1}{|\mathcal{B}_{\text{pred}}|} \sum_{b_{\text{pred}} \in \mathcal{B}_{\text{pred}}} \mathbf{1}_{\left( \max_{b_{\text{gt}} \in \mathcal{B}_{\text{gt}}} \text{IoU}(b_{\text{pred}}, b_{\text{gt}}) \ge 0.5 \right)} 
\]
Precision is defined as $0$ when the model produces no bounding box ($|\mathcal{B}_{\text{pred}}| = \emptyset$).

\textbf{F1-Score (F1)} To balance the trade-off between localization recall and precision, we report the $F_1$ score, which is the harmonic mean of the standard Recall ($\text{Rec.}$) defined in the main text and the Precision ($\text{Prec.}$) defined above:
\[ 
F_1 = 2 \cdot \frac{\text{Prec.} \cdot \text{Rec.}}{\text{Prec.} + \text{Rec.}} 
\]
This metric provides a single scalar value to evaluate the overall efficiency of the evidence extraction process, ensuring the model is both thorough and concise in its attribution. Note that F1 is computed per sample and then averaged over the dataset, rather than from the aggregated Precision and Recall.

\begin{table}[htbp]
\centering
\scriptsize
\renewcommand{\arraystretch}{1.25} 
\setlength{\tabcolsep}{1.0pt} 
\caption{Detailed attribution performance of MLLMs across different document scenarios. Page. denotes Page-level Recall; Rec., Prec., and F1 represent bounding-box-level Recall, Precision, and F1-score respectively. For each metric, the best and second-best results are highlighted in \highlightblue{blue} and \highlightgreen{green}, respectively.}
\label{tab:multi_scenario_results}

\begin{tabular}{@{}l cccc cccc cccc >{\bfseries}c>{\bfseries}c>{\bfseries}c>{\bfseries}c@{}}
\toprule
\multirow{2.5}{*}{\textbf{Model}} & \multicolumn{4}{c}{\textbf{Single-Doc}} & \multicolumn{4}{c}{\textbf{Multi (1-Gold)}} & \multicolumn{4}{c}{\textbf{Multi (N-Gold)}} & \multicolumn{4}{c}{\textbf{Overall}} \\
\cmidrule(lr){2-5} \cmidrule(lr){6-9} \cmidrule(lr){10-13} \cmidrule(l){14-17}
& Page. & Rec. & Prec. & F1 & Page. & Rec. & Prec. & F1 & Page. & Rec. & Prec. & F1 & Page. & Rec. & Prec. & F1 \\ \midrule
% 分类1：闭源
\multicolumn{17}{c}{\textit{Closed-source MLLMs}} \\ \midrule
Gemini-3.1-Pro-Preview &88.8&\highlightblue{68.9}&\highlightblue{63.1}&\highlightblue{62.0}&91.5&\highlightblue{69.4}&\highlightblue{62.8}&\highlightblue{61.6}&81.4&\highlightblue{55.3}&\highlightblue{49.4}&\highlightblue{48.6}&87.9&\highlightblue{66.0}&\highlightblue{59.9}&\highlightblue{58.9} \\

Gemini-3-Flash-Preview &\highlightblue{92.8}&\highlightgreen{49.5}&\highlightgreen{37.0}&\highlightgreen{37.9}&\highlightblue{94.3}&\highlightgreen{42.1}&\highlightgreen{30.1}&\highlightgreen{31.3}&\highlightblue{86.5}&\highlightgreen{39.5}&\highlightgreen{29.9}&\highlightgreen{30.3}&\highlightblue{91.8}&\highlightgreen{45.4}&\highlightgreen{33.7}&\highlightgreen{34.5} \\ 

Gemini-2.5-Pro &\highlightgreen{92.3}&31.5&24.8&24.6&\highlightgreen{93.5}&25.4&18.3&18.6&\highlightgreen{81.7}&20.0&17.4&16.2&\highlightgreen{90.3}&27.4&21.5&21.2\\

GPT-5.4&88.5&35.9&29.6&29.6&79.9&25.7&22.4&20.9&75.4&25.7&21.2&20.6&83.4&31.0&25.9&25.4\\ 

GPT-5.2 &67.4&20.9&20.4&18.6&75.9&16.5&15.4&14.8&66.1&13.9&13.4&12.0&69.3&18.2&17.6&16.2\\

Qwen3.6-Plus&50.0&9.8&9.7&8.7&51.7&5.9&6.0&5.5&47.5&4.6&4.9&4.3&49.9&7.7&7.7&6.9\\

Seed2.0-Pro&69.7&35.8&31.8&31.2&59.5&18.1&16.9&15.5&56.4&21.5&18.7&17.2&64.4&28.5&25.4&24.4\\

GLM-5V-Turbo&48.9&18.3&18.5&16.4&43.3&11.7&12.4&10.8&44.0&10.2&9.3&9.0&46.5&14.9&15.0&13.4 \\
\midrule

% 分类2：开源大模型
\multicolumn{17}{c}{\textit{Open-source Large MLLMs}} \\ \midrule
Kimi-K2.5 &44.8&8.2&8.0&7.4&42.4&3.5&3.2&3.0&46.0&4.8&5.7&4.4&44.4&6.2&6.3&5.6  \\ 
Gemma-4-31B &44.8&10.9&9.8&9.0&56.0&14.0&12.4&11.0&51.1&10.4&11.0&9.0&49.1&11.6&10.7&9.5\\ 
Qwen3.5-397B-A17B&48.8&6.8&6.9&6.1&41.6&4.0&3.3&3.4&41.7&3.8&4.5&3.6&45.3&5.4&5.4&4.9\\
Qwen3.5-122B-A10B&41.5&5.9&5.4&5.2&28.6&1.7&1.7&1.6&30.1&1.9&2.1&1.6&35.6&3.9&3.7&3.5 \\
Qwen3.5-27B &50.2&7.0&6.9&6.0&43.4&3.1&2.4&2.3&45.1&3.9&3.7&3.5&47.3&5.3&5.0&4.5  \\
Qwen3-VL-235B-A22B &64.4&15.2&14.9&13.5&50.7&6.2&5.7&5.4&50.5&8.1&7.5&7.0&57.8&11.3&10.9&10.0\\
Qwen3-VL-32B&69.0&8.0&7.5&6.8&50.7&2.8&2.7&2.4&52.9&7.9&7.5&7.0&60.7&6.6&6.3&5.7\\ 
\midrule

% 分类3：开源小模型
\multicolumn{17}{c}{\textit{Open-source Small MLLMs}} \\ \midrule
Gemma-4-26B-A4B &20.8&2.2&2.7&1.9&22.0&4.2&5.2&3.7&27.5&3.5&3.9&3.2&22.6&3.0&3.6&2.7 \\ 
Qwen3.5-35B-A3B &35.1&2.7&2.4&2.3&12.4&0.5&0.5&0.5&18.5&0.6&1.0&0.6&25.5&1.7&1.6&1.5 \\
Qwen3.5-9B &33.6&2.5&2.4&2.3&6.2&0.3&0.3&0.3&12.2&0.8&1.3&0.9&21.8&1.6&1.6&1.5
 \\
Qwen3-VL-30B-A3B  &23.0&5.6&6.1&5.0&7.1&0.9&0.9&0.9&14.1&1.7&2.4&1.6&16.9&3.5&4.0&3.2\\
Qwen3-VL-8B   &43.2&1.8&1.5&1.4&18.9&0.0&0.0&0.0&16.9&0.3&0.2&0.3&31.1&1.0&0.9&0.8
   \\
\bottomrule
\end{tabular}
\end{table}

\subsection{More Results of Experiments}
\label{Appendix: More Results of Experiments}
\paragraph{Widespread Deficiency in Coarse-grained Attribution} A striking observation from Table~\ref{tab:multi_scenario_results} is that Page-level Recall (Page.) remains remarkably low for the vast majority of models. This indicates that the failure in evidence attribution is not merely a consequence of weak fine-grained grounding (i.e., missing the exact box), but a more fundamental inability to navigate to the correct document page. While the Gemini-3 series demonstrates strong navigation (above 87\% Overall Page.), many advanced models like GPT-5.2 (69.3\% Overall Page.) and Qwen3-VL-235B-A22B (57.8\% Overall Page.) frequently fail to even locate the relevant page. This "coarse-level blindness" suggests that before addressing spatial precision, models must first overcome significant hurdles in document-level retrieval and page indexing.

\paragraph{Impact of Multi-document Complexity} The challenge of attribution is significantly exacerbated as the environment shifts from Single-Doc to Multi-Doc (N-Gold) settings. In these high-density scenarios, even top-tier models exhibit a sharp performance collapse. For instance, GPT-5.4 sees its Page-level Recall drop from 88.5\% in Single-Doc to 75.4\% in N-Gold, while its F1-score falls from 29.6\% to 20.6\%. For many open-source models, the Multi-Doc setting acts as a performance ceiling; for example, Qwen3-VL-235B-A22B experiences a significant decline in Page-level Recall, dropping from 64.4\% to 50.5\%. This trend underscores that current MLLMs lack the robustness required for multi-document reasoning, a deficit that directly fuels the observed "Attribution Hallucination."

\subsection{More Fine-grained Results}
\label{Appendix: Fine-grained Results}
\begin{figure}[htbp]
\centering
\includegraphics[width=0.5\textwidth]{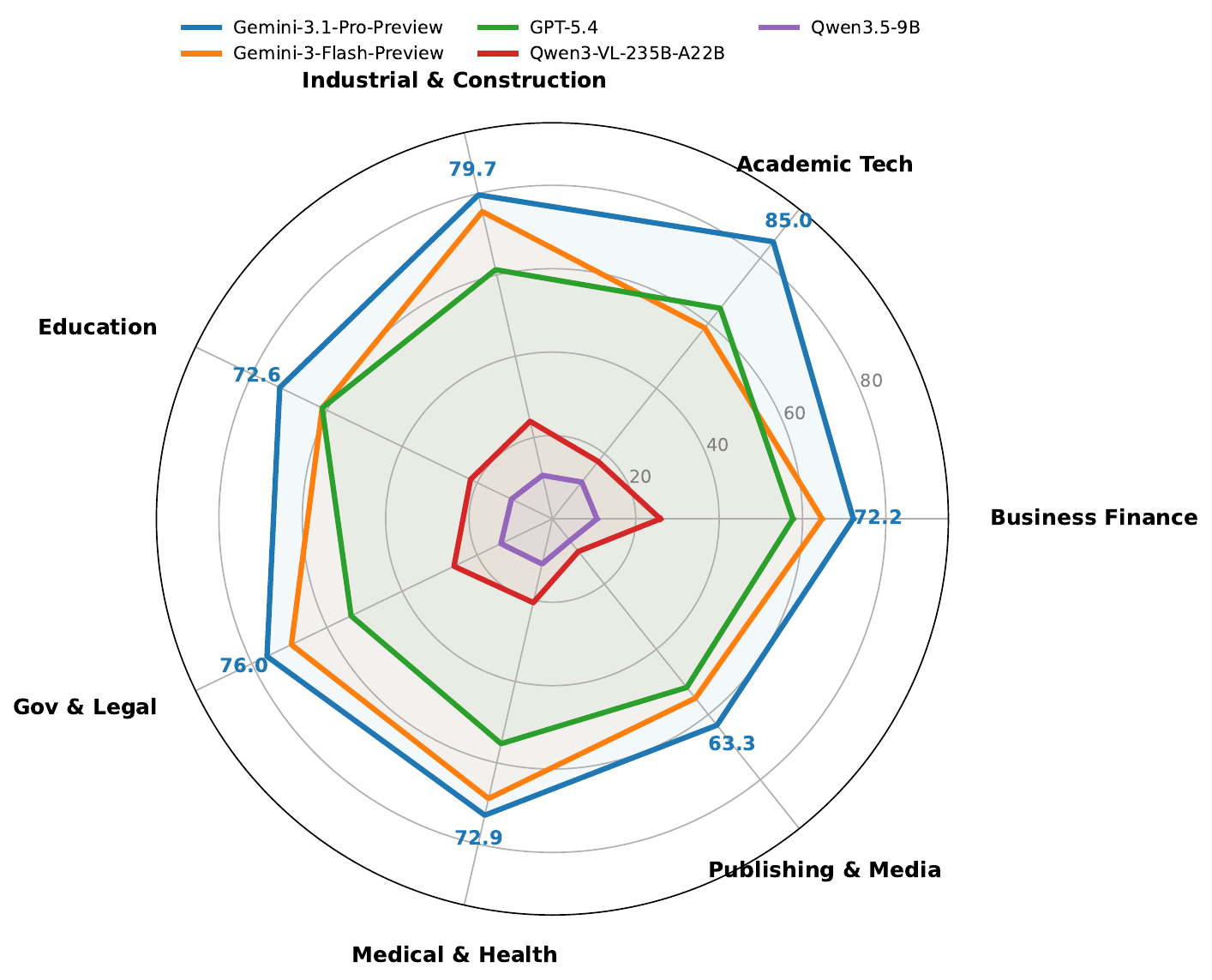}
\caption{Fine-grained results on various document types (SAA Score).}
\label{fig:citevqa_domain_radar}
\end{figure}

\paragraph{Document Type} Our cross-domain evaluation reveals that performance varies significantly depending on document structure (See Figure~\ref{fig:citevqa_domain_radar}): models achieve peak performance in the Academic Tech domain (e.g., Gemini-3.1-Pro-Preview at 85.0) due to the highly standardized layouts and logical rigor of academic papers, which facilitate precise evidence attribution. Conversely, the Publishing \& Media domain presents the greatest challenge, with the highest SAA reaching only 63.3, as the complex typographic designs, non-linear content distribution, and intricate image-text interleaving inherent in newspapers and magazines severely hinder models' spatial perception and cross-page reasoning capabilities.

\begin{figure}[htbp]
\centering
\includegraphics[width=0.5\textwidth]{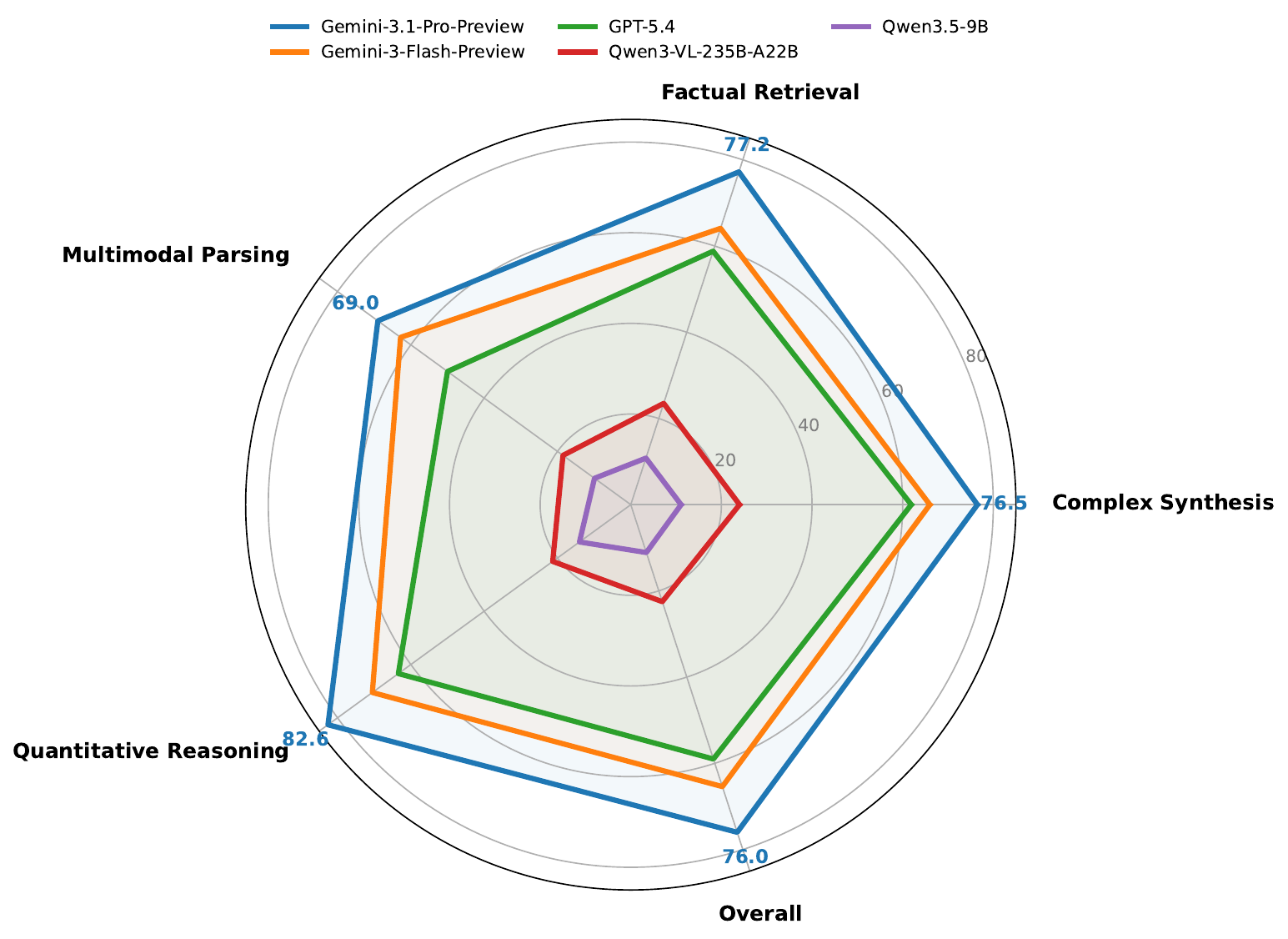}
\caption{Fine-grained results on various question types (SAA Score).}
\label{fig:citevqa_ability_radar}
\end{figure}

\paragraph{Question Type} Results show a significant performance gap between question types (See Figure~\ref{fig:citevqa_ability_radar}). Models excel in Quantitative Reasoning (e.g., Gemini-3.1-Pro-Preview at 82.6) because numerical computations rely on objective logic and offer clear alignment between evidence and answers. In contrast, the newly introduced Multimodal Parsing task remains a major bottleneck; this category requires models to locate specific document elements based on descriptive cues (such as identifying a particular table by its background color or section header) and subsequently parse the content, leading to substantial difficulties in both precise evidence attribution and final answer generation. 

\subsection{Details of SAA Failure Decomposition}
\label{Appendix: Error Decomposition}

To quantify the failure modes behind SAA$=0$, we classify every failed sample into four mutually exclusive categories. We first define a lenient overlap measure, min-area-coverage:
\[
\mathrm{MinCov}(b_1, b_2) = \frac{\mathrm{Area}(b_1 \cap b_2)}{\min(\mathrm{Area}(b_1), \mathrm{Area}(b_2))},
\]
which is high whenever one box (nearly) contains the other, regardless of size mismatch. Both scores are computed against the \emph{crucial} evidence only and aggregated at the sample level via the best-matching pair: $\mathrm{IoU} = \max_{b_{\text{pred}} \in \mathcal{B}_{\text{pred}},\, b_{\text{gt}} \in \mathcal{B}_{\text{crucial}}} \mathrm{IoU}(b_{\text{pred}}, b_{\text{gt}})$, and analogously for MinCov---consistent with Section~4.1, where Recall also matches predictions to crucial boxes by maximum IoU. The categories are:
\begin{itemize}
    \item \textbf{Cat1 (Answer Wrong)}: Ans.\ $< 4$.
    \item \textbf{Cat2 (Page Wrong)}: Ans.\ $\geq 4$, but the predicted evidence lies on a wrong page, or no bounding box is produced.
    \item \textbf{Cat3 (Wrong Element)}: Ans.\ $\geq 4$, correct page, but IoU $< 0.3$ \emph{and} MinCov $< 0.5$---the prediction has essentially no spatial relation to any crucial element.
    \item \textbf{Cat4 (Boundary Imprecision)}: Ans.\ $\geq 4$, correct page, and IoU $\in [0.3, 0.5)$ \emph{or} MinCov $\geq 0.5$---a crucial element is found but the box is imprecise.
\end{itemize}
Note how low the bar for escaping Cat3 already is: a prediction avoids Cat3 whenever \emph{either} IoU $\geq 0.3$ \emph{or} MinCov $\geq 0.5$ holds, and MinCov is inherently lenient---it scores high as long as one box merely contains the other. A prediction failing both conditions therefore shares essentially no area with any crucial element, so Cat3 reflects genuinely wrong-element citation rather than borderline imprecision. Figure~\ref{fig:error_decomposition} reports the resulting breakdown (percentages within each model's SAA$=0$ samples).

\clearpage
\newpage
\section{Case Study}
\label{Appendix: Case Study}
\begin{figure}[htbp]
\centering
\includegraphics[width=0.8\textwidth]{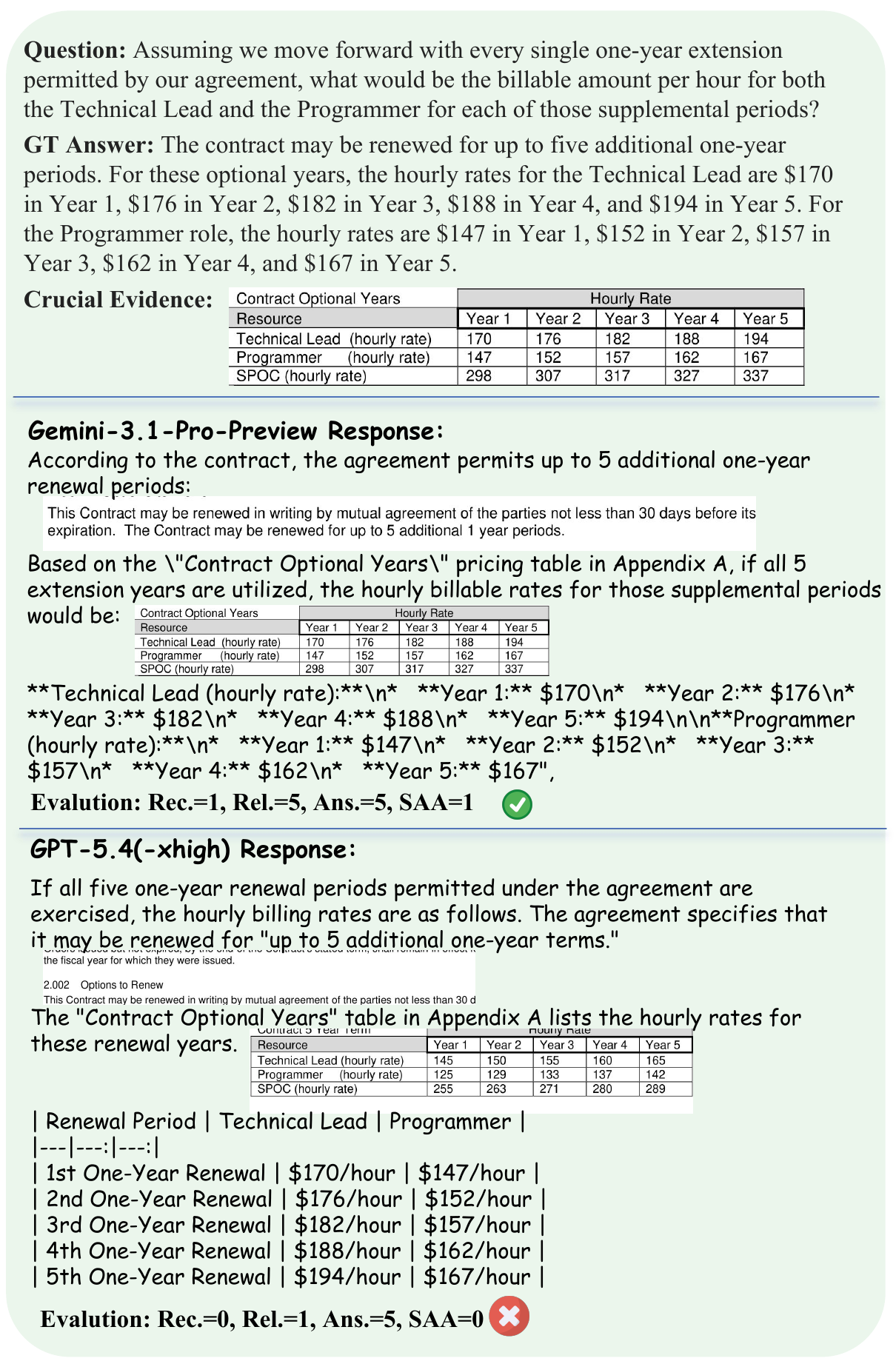}
\caption{Case Study 1. While both models generate correct answers (Ans.=5), Gemini-3.1-Pro-Preview accurately cites the "Contract Optional Years" table (SAA=1). Conversely, GPT-5.4 exhibits "Attribution Hallucination" by providing the correct text but citing an incorrect pricing table (e.g., \$145 vs. \$170 for Year 1), resulting in Rec.=0 and SAA=0.}
\label{fig:Case_study1}
\end{figure}

\begin{figure}[htbp]
\centering
\includegraphics[width=0.8\textwidth]{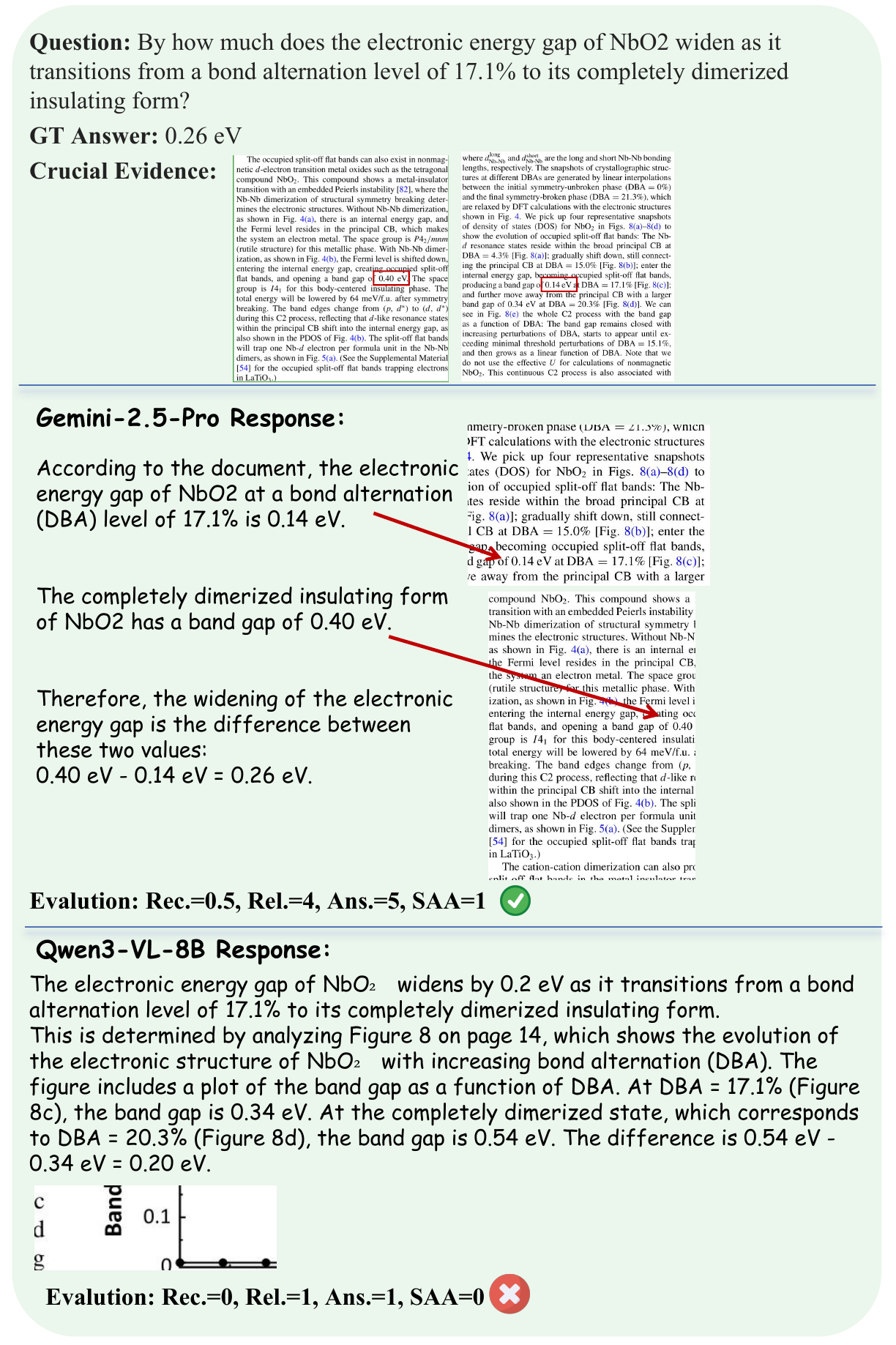}
\caption{Case Study 2. While Gemini-2.5-Pro correctly calculates the widening gap ($0.40 - 0.14 = 0.26$ eV) and cites the corresponding evidence segments (SAA=1), Qwen3-VL-8B fails both semantically and visually. It extracts incorrect values ($0.34$ and $0.54$ eV) from the text and provides irrelevant citations, resulting in Ans.=1 and SAA=0. This case demonstrates that even when evidence is explicitly stated in the text, weaker models struggle with both the retrieval and the logic required for multi-step attribution. }
\label{fig:Case_study2}
\end{figure}

\newpage
\section{Prompt Templates}
\label{Prompt Templates}
\subsection{Prompts for CiteVQA Pipeline}
\label{Prompts for CiteVQA Pipeline}

\begin{promptbox}[breakable]{Prompt for Extracting Evidence Package from PDFs}
**Role**: Parse a PDF document containing outline, OCR text blocks, bounding boxes, and page screenshots.

**Goal**: Collect high-quality, verifiable evidence bundles to support Q&A, analysis, calculation, and visual extraction.

**Each evidence bundle MUST satisfy**:

1. **Multi-page**: At least 2 different pages.
2. **Multi-element**: At least 2 element types (e.g., text, table, figure, layout).
3. **Complete context**:
   - If including a table/figure, MUST also extract its title, caption, legend, axis labels, footnotes, etc.
   - If an element spans multiple pages (e.g., continued table), MUST extract the complete structure from ALL involved pages (e.g., headers from previous page).

**What to capture**:
- **Text**: Key phrases, definitions, scope notes.
- **Figures**: Captions, axis/legend text, panel labels.
- **Tables**: Full headers, target cells, footnotes.
- **Layout**: Relative position, grouping, visual prominence (e.g., full-width table, top large figure).

**Exploration steps**:
1. Search keywords: "Figure", "Table", "Note", "unit", etc.
2. Extract the hit AND all surrounding relevant elements (enforce complete context for tables/figures).
3. Cross-page link: Connect same metric/entity across different pages.
4. Use screenshots and bounding boxes to confirm type and layout.

**Avoid**:
- Single-element bundles, ID/page-number dependencies, fragmented tables/figures, broad summaries without clear Q&A target.

**Output format**: Return a list of evidence bundles. Generate at least **10** bundles.

```json
[{
  "Evidence_package_description": "Brief description of purpose and reasoning value",
  "Evidence_list": [
    {
      "type": "element_type",
      "content": "OCR text content",
      "bbox": [y1, x1, y2, x2],
      "angle": 0,
      "page_id": page_number,
      "element_idx": "element_index"
    }
    // At least 5 relevant elements
  ]
}]
```
\end{promptbox}

\begin{promptbox}[breakable]{Prompt for Getting Templates from Open-source Datasets}
Given question samples, generate one reusable template for each sample.

Return strict JSON only:
{{
  "samples": [
    {{
      "sample_id": 0,
      "category": "...",
      "template_en": "...",
      "template_cn": "...",
      "example_en": "...",
      "example_cn": "..."
    }}
  ]
}}

Requirements:
1. category: "Complex Synthesis", "Factual Retrieval", "Multimodal Parsing", "Quantitative Reasoning",.
2. template_en/template_cn must be abstract reusable templates, use placeholders like [Entity], [Date], [Metric], [Section], [Method].
3. example_en/example_cn should be one short concrete question in that sample style.
4. Keep semantic consistency within each sample.
5. Must return one item for every sample_id provided.
\end{promptbox}

\begin{promptbox}[breakable]{Prompt for Annotation Evaluation}

You are an expert evaluator for a VQA benchmark. Your task is to assess the quality of a given QA pair along three dimensions: **Question Difficulty**, **Answer Quality**, and **Crucial Evidence Quality**.

Please follow the scoring criteria below. All scores range from **0 to 5**, where 0 indicates complete failure or unusable quality.

---

**1. Question Difficulty (0-5)**
- **0**: Nonsensical, unanswerable, or no meaningful question.
- **1 to 2 (Simple)**: Direct fact retrieval, minimal reasoning, no cross-document or cross-page synthesis.
- **3 (Moderate)**: Requires basic inference or aggregation from a single document section.
- **4 to 5 (Complex)**: Involves multi-step reasoning, cross-document comparison, contradiction resolution, or indirect evidence extraction.

---

**2. Answer Quality (0-5)**
- **0**: No answer or completely irrelevant.
- **1 to 2 (Poor)**: Largely incorrect, missing key information, or no citation.
- **3 (Acceptable)**: Correct but overly brief, lacks sufficient justification or citation.
- **4 to 5 (Good/Excellent)**: Accurate, well-structured, properly cited, and fully addresses the question.

---

**3. Crucial Evidence Quality (0-5)**
- **0**: No evidence provided, or evidence completely unrelated.
- **1 to 2 (Weak)**: Evidence is minimally relevant or insufficient to support the answer.
- **3 (Moderate)**: Evidence is relevant but incomplete, not optimally cited, or overly redundant.
- **4 to 5 (Strong)**: Evidence precisely supports the answer, comes from authoritative sections (e.g., tables, core arguments), and includes necessary span-level references.

---

**Output Format:**

Please output in the following structured format:

```
Question Difficulty: [0 to 5]
Brief Justification: ...

Answer Quality: [0 to 5]
Brief Justification: ...

Evidence Quality: [0 to 5]
Brief Justification: ...
```
\end{promptbox}

\newpage
\subsection{Prompts for CiteVQA Evaluation}
\label{Prompts for CiteVQA Evaluation}
\begin{promptbox}[breakable]{Prompt for Inference in Single-Doc}
# Document Analysis Assistant

Answer the question based on the provided PDF page images, and cite the evidence regions in your answer.

## Evidence Citation Rules

1. Evidence must be at the **element level**: a complete paragraph, a complete table, a complete image, or a complete note. Do not select partial text from a paragraph or a single row from a table, and do not select an entire page or spanning multiple tables/paragraphs. Note: This is very important and will directly affect your score.
2. For **tables and images**, if there are captions or footnotes, they need to be annotated as **separate evidence** with their own bbox, not merged into the table/image bbox.
3. Each piece of cited evidence text should be followed by a `<bbox />` tag indicating the evidence location.
4. When an inference step relies on multiple pieces of evidence, use multiple `<bbox />` tags separately.
5. Pure reasoning/calculation steps do not need `<bbox />`.

## Annotation Format

```
<bbox page="page_number" x1="left" y1="top" x2="right" y2="bottom" />
```

Page numbers start from 1 (note: ignore original page numbers); coordinates are relative coordinates on the page image, range 0-1000.

## Examples

**Question:** What is the net change in the company's precision copper tube production capacity from 2021 to 2024?

**Answer:**

According to the main text, the company's precision copper tube production capacity increased from 798,000 tons in 2021 to 1.31 million tons in 2024:
<bbox page="1" x1="536" y1="65" x2="642" y2="656" />

Therefore, the net change = 1.31 - 0.798 = 0.512 million tons.

Additionally, according to "Table 1: Production line renovation will reduce the company's costs", the per-ton comprehensive cost is expected to decrease by 700 yuan/ton after the production line renovation:
<bbox page="8" x1="584" y1="65" x2="598" y2="371" />
<bbox page="8" x1="598" y1="59" x2="712" y2="670" />

## Final Reminder
Evidence must be a complete paragraph, a complete table, a complete image, or a complete note. Do not select partial rows from a paragraph or a single row from a table, and do not select an entire page or spanning multiple tables/paragraphs. Note: This is very important and will directly affect your score.
\end{promptbox}

\begin{promptbox}[breakable]{Prompt for Inference in Multi-Doc}
# Multi-Document Analysis Assistant

Answer the question based on the provided PDF documents, and cite the evidence regions in your answer.

## Document Numbering Rules

- Document numbering starts from 1, corresponding to the order in the `PDF_Source` list
- Must use the correct document numbers when citing evidence

## Evidence Citation Rules

1. Evidence must be at the **element level**: a complete paragraph, a complete table, a complete image, or a complete note. Do not select partial text from a paragraph or a single row from a table, and do not select an entire page or spanning multiple tables/paragraphs. Note: This is very important and will directly affect your score.
2. For **tables and images**, if there are captions or footnotes, they need to be annotated as **separate evidence** with their own bbox, not merged into the table/image bbox.
3. Each piece of cited evidence text should be followed by a `<bbox />` tag indicating the evidence location.
4. When an inference step relies on multiple pieces of evidence, use multiple `<bbox />` tags separately.
5. Pure reasoning/calculation steps do not need `<bbox />`.

## Annotation Format
```
<bbox doc="document_number" page="page_number" x1="left" y1="top" x2="right" y2="bottom" />
```
- `doc`: Document number, starting from 1 (corresponding to `PDF_Source` list order)
- `page`: Page number starting from 1 (note: ignore original page numbers)
- Coordinates are relative coordinates on the page image, range 0-1000

## Examples
**Question:** Compare the revenue data differences between Company in Document 1 and Document 2.

**Answer:**
According to the main text of Document 1, the company's 2023 revenue was 10 billion yuan:
<bbox doc="1" page="1" x1="536" y1="65" x2="642" y2="656" />

According to the financial report in Document 2, the company's 2023 revenue was 12 billion yuan:
<bbox doc="2" page="3" x1="584" y1="65" x2="598" y2="371" />

Therefore, the revenue difference reported in the two documents is 2 billion yuan.

## Final Reminder
Evidence must be a complete paragraph, a complete table, a complete image, or a complete note. Do not select partial rows from a paragraph or a single row from a table, and do not select an entire page or spanning multiple tables/paragraphs. Note: This is very important and will directly affect your score.
\end{promptbox}

\begin{promptbox}[breakable]{Prompt for Evaluating Relevance}
## Task
You are a professional DocVQA quality evaluation expert. Your task is to evaluate whether the PDF screenshots referenced in the answer can effectively support the corresponding answer content. You need to determine whether the visual information (text, charts, data) in the screenshots is consistent with the facts mentioned in the answer.

You will receive a question, a standard answer (without images), and the model's generated answer with interleaved images.

## Evaluation Dimensions
- Truthfulness: Does the screenshot contain the key data or descriptions mentioned in the answer?
- Sufficiency: Does the screenshot provide sufficient evidence for the conclusion, or is it taken out of context?
- Localization Accuracy: Does the screenshot precisely cover the answer source, or does it contain irrelevant information?
- Alignment: Does the screenshot exactly match the text being cited? Any misalignment is a flaw.

## Scoring Criteria
**BE STRICT. A score of 5 is extremely rare and requires perfection. Most good answers should score 3-4.**
- 0: No support at all. The screenshot content is completely irrelevant to the answer.
- 1: Extremely weak support. The screenshot only mentions vague related concepts without specific data.
- 2: Weak support. The screenshot contains partial key data, or has significant quality issues.
- 3: Moderate support. The screenshot covers most of the evidence but has flaws (e.g., includes irrelevant content, slight misalignment with cited text, or captures too much/too little).
- 4: Good support. The screenshot contains the core evidence with minor flaws. This is where most correct answers should score.
- 5: **PERFECT support (extremely rare)**. The screenshot must be **flawless**: precise bounding box that exactly covers the cited text, no extra content, no skewing, no misalignment, and the evidence perfectly matches what is claimed. **Only give 5 when every single detail is perfect.**

## Important Notes
- Be conservative with scores. If you hesitate between two scores, choose the lower one.
- A slightly off-center crop, a small amount of extra content, or minor misalignment = score 3-4, NOT 5.
- Score 5 should only be given when the bounding box is pixel-perfect and the evidence is exactly what was cited.

## Output Format    
Please output two lines for the results: the first line is your reasoning for the score, and the second line is the score. Strictly follow this format without any additional content.

# Output Example  
A reason why you choose this score (from 0 to 5).
```<relevance_score>X</relevance_score>```
\end{promptbox}

\begin{promptbox}[breakable]{Prompt for Evaluating Answer Correctness}
## Task
You are a multimodal QA evaluation expert. Your task is to evaluate the overall quality of the answer. Provide your evaluation in the form of "reasoning" and "score". Evaluation should be based solely on the standard answer, without introducing your own external knowledge.
You will receive a question, a standard answer, and the model's generated answer.

## Evaluation Criteria
**BE STRICT. Most answers are not as good as they appear. When in doubt, choose the lower score.**
- 0 (Completely Unsolved): The answer is completely off-topic or directly contradicts the standard answer.
- 1 (Mostly Unsolved): The answer has extremely low relevance, providing almost no valuable information.
- 2 (Partially Solved): The answer covers some aspects but misses key information or has obvious factual errors. **Many "okay" answers fall here - do not over-rate.**
- 3 (Acceptable): The answer covers the core facts but is incomplete, lacks necessary details, or has minor errors. **Only give this when the answer is genuinely useful despite clear gaps.**
- 4 (Good): The answer clearly covers all key points with rigorous logic. Near-complete and accurate. **Reserve for strong answers. Do not hand out freely.**
- 5 (Excellent): Complete, accurate, and perfectly structured and the answer must not be significantly more verbose than the standard answer.  **Extremely Difficult to reach. Do not give 5 unless truly perfect in every dimension.**

## Important Notes
- Ignore phrases like "cited from" or "from" that may appear in the model's generated answer - they are irrelevant.
- **DO NOT penalize the answer based on the language it is written in.** Chinese, English, or mixed - score the content only.
- Only the exact facts in the standard answer count. Extra details beyond the standard answer do NOT improve the score.

## Output Format    
Please output two lines for the results: the first line is your reasoning for the score, and the second line is the score. Strictly follow this format without any additional content.

# Output Example  
A reason why you choose this score (from 0 to 5).
```<qa_acc>X</qa_acc>```

\end{promptbox}

\clearpage
\newpage
\section{Limitations}
\label{appenidx: Limitations}
While CiteVQA introduces a rigorous framework for traceable document intelligence, it entails certain inherent trade-offs. First, although the benchmark spans seven major domains, the definition of authoritative evidence may involve domain-specific nuances in highly specialized vertical fields that warrant further exploration. Second, our automated curation pipeline prioritizes data fidelity by leveraging state-of-the-art Multimodal Large Language Models (MLLMs), which, while ensuring high-quality reasoning and attribution, introduces a significant computational resource barrier for large-scale replication. Third, the multi-dimensional evaluation protocol—incorporating coordinate verification and fine-grained textual alignment—requires higher computational overhead compared to standard VQA tasks, representing a deliberate choice to prioritize evaluative depth and traceability over raw scoring efficiency. Finally, since the source PDFs are publicly available, some documents may have been seen during model pretraining or post-training data collection; part of the measured performance may thus reflect prior exposure rather than purely evidence-grounded reasoning over unseen inputs. We plan a private held-out subset sourced from post-cutoff documents for contamination-free evaluation in future work. Additionally, our evaluation protocol necessarily adapts visual inputs to each model's interface and context window---native document APIs for some proprietary systems and resolution-scaled page images for others---so cross-model comparisons should be interpreted with these input-fidelity differences in mind.

%paragraph{Potential  Negative Impacts} A potential negative impact is the risk of models overfitting to the specific metrics and document distributions of CiteVQA. While our benchmark aim to improve document intelligence, excessive optimization for these specific tasks may lead to reduced generalizability when models encounter diverse real-world document structures not represented in our dataset.  
% ICLR has no paper checklist, so the NeurIPS checklist is not included here.
% The original file is kept as checklist_neurips_unused.tex for reference; it
% relies on \answerYes/\answerNo macros that only exist in the NeurIPS style.
% \input{checklist.tex}

\end{document}